\documentclass[letterpaper]{article} 
\usepackage{aaai2026}
\usepackage{times}  
\usepackage{helvet}  
\usepackage{courier}  
\usepackage[hyphens]{url}  
\usepackage{graphicx} 
\usepackage{natbib}  
\usepackage{caption} 
\usepackage{algorithm, algpseudocode}

\usepackage{newfloat}
\usepackage{listings}
\DeclareCaptionStyle{ruled}{labelfont=normalfont,labelsep=colon,strut=off} 
\floatstyle{ruled}
\newfloat{listing}{tb}{lst}{}
\floatname{listing}{Listing}
\def\rvx{{\mathbf{x}}}
\newcommand{\eqn}[1]{
    \begin{align}{#1}\end{align}
}
\def\Ls{\mathcal{L}}
\def\E{\mathbb{E}}
\usepackage{subcaption}

\usepackage{adjustbox}
\usepackage[font=small,labelfont=bf]{caption}
\usepackage{multirow}
\usepackage{amsmath}
\usepackage{amssymb}
\usepackage{mathtools}
\usepackage{amsthm}
\usepackage{adjustbox}
\usepackage{booktabs} 

\usepackage{copy_defs_ahn}

\title{Extendable Planning via Multiscale Diffusion}

\author {
    Chang Chen\equalcontrib\textsuperscript{\rm 2},
    Hany Hamed\equalcontrib\textsuperscript{\rm 1,3},
    Doojin Baek\textsuperscript{\rm 1},
    Taegu Kang\textsuperscript{\rm 1},
    Samyeul Noh\textsuperscript{\rm 4},\\
    Yoshua Bengio\textsuperscript{\rm 5},
    Sungjin Ahn\textsuperscript{\rm 1,6}
}
\affiliations {
    \textsuperscript{\rm 1}KAIST, 
    \textsuperscript{\rm 2}Rutgers University, 
    \textsuperscript{\rm 3}IIT,  
    \textsuperscript{\rm 4}ETRI, 
    \textsuperscript{\rm 5}Mila, 
    \textsuperscript{\rm 6}NYU
}

\begin{document}

\maketitle

\begin{abstract}
Long-horizon planning is crucial in complex environments, but diffusion-based planners like Diffuser are limited by the trajectory lengths observed during training. This creates a dilemma: long trajectories are needed for effective planning, yet they degrade model performance. In this paper, we introduce this \textit{extendable long-horizon planning} challenge and propose a two-phase solution. First, Progressive Trajectory Extension incrementally constructs longer trajectories through multi-round compositional stitching. Second, the Hierarchical Multiscale Diffuser enables efficient training and inference over long horizons by reasoning across temporal scales. To avoid the need for multiple separate models, we propose Adaptive Plan Pondering and the Recursive HM-Diffuser, which unify hierarchical planning within a single model. Experiments show our approach yields strong performance gains, advancing scalable and efficient decision-making over long-horizons.
\end{abstract}

\section{Introduction}\label{sec:intro} 
The ability to plan over long horizons via a learned world model ~\citep{hamrick2020role,mattar2022planning} enables agents to pursue long-term goals, even in environments with sparse rewards~\citep{alphago,hafner2019learning,Hansen2022tdmpc}. However, constructing effective world models~\citep{worldmodels} for such planning remains a challenge. Traditional planning approaches based on autoregressive forward models are particularly susceptible to compounding errors~\citep{lambert2022investigating,bachmann2024pitfalls}.
In offline reinforcement learning, the Diffuser framework~\citep{janner2022diffuser,decision_diffuser} offers a promising alternative by leveraging diffusion models~\citep{sohl2015deep,ho2020denoising} to bypass explicit forward dynamics. Instead of predicting states sequentially, Diffuser generates entire trajectories holistically thereby mitigating compounding errors and improving planning accuracy, particularly for long-horizon planning.

A fundamental but underappreciated limitation of applying diffusion models to long-horizon planning is that they cannot plan beyond the trajectory lengths seen during training. As a result, modeling trajectories longer than those seen during training becomes difficult. However, many real-world applications require the ability to plan beyond the observed sequence lengths. In contrast, forward model-based planning does not suffer from this problem as it can extend to previously unseen horizons by rolling out longer sequences. Therefore, it is crucial for diffusion-based planner to address this issue.

While one solution is to collect longer trajectories, this quickly becomes impractical. For example, enabling a robot to plan over week- or month-long horizons would require uninterrupted trajectories of that length—an increasingly infeasible task as the horizon grows. Even if such data were available, training a Diffuser model on trajectories of that scale remains challenging. In fact, planning performance has been shown to degrade on extended sequences in existing diffusion-based approaches~\citep{hd}. This presents a fundamental dilemma: \textit{long trajectories are needed for effective planning, yet they undermine the performance of the Diffuser itself}.

In this paper, we propose a simple two-phase solution to this fundamental dilemma, which we term the \textit{extendable long-horizon planning} challenge. To address this problem, we propose both a mechanism to generate long trajectories from short ones and a model capable of learning effectively from them.

First, we introduce Progressive Trajectory Extension (PTE)—a method that incrementally extends short training trajectories into significantly longer ones. While trajectory stitching is not a new concept, PTE differs from traditional approaches by not merely augmenting data, but by actively extending trajectories through multiple rounds of compositional stitching of previously extended sequences. This multi-round process enables the construction of extremely long trajectories. 

Second, we present the Hierarchical Multiscale Diffuser (HM-Diffuser), which enables diffusion models to be trained effectively on these long trajectories by introducing multiple temporal scales. In particular, to overcome limitations of existing hierarchical Diffuser frameworks—which require separate models for each hierarchy level—we further propose the Adaptive Plan Pondering and the Recursive HM-Diffuser in the HM-Diffuser framework. These techniques consolidate multiple hierarchical layers into a single model capable of recursively reasoning across temporal scales.

Our experiments demonstrate that the integration of PTE and HM-Diffuser yields non-trivial benefits. The combined approach significantly enhances performance across a range of planning tasks, underscoring its potential to advance scalable and efficient long-horizon decision-making.

The key contributions of this paper are as follows. First and foremost, this paper identifies and formulates a previously underexplored challenge: extendable long-horizon planning in Diffuser-based models. Second, as a solution to this novel challenge, we propose a new unified two-phase framework that integrates a process for generating extended trajectories from short demonstrations, and a learning framework that can leverage them effectively. While each component draws from prior work on trajectory stitching and hierarchical diffusion, their integration to tackle this specific challenge is novel. Third, we present PTE, a multi-round stitching method that generates explicitly longer trajectories through iterative compositional extension. Fourth, we introduce the Recursive Hierarchical Diffuser, a model that enables efficient long-horizon planning through multiscale hierarchical reasoning. Finally, we introduce the Plan Extendable Trajectory Suite (PETS) benchmark.

\section{Preliminaries}

\textbf{Diffusion models}~\citep{sohl2015deep,ho2020denoising}, inspired by the modeling of diffusion processes in statistical physics, are latent variable models with the following generative process 
\vspace{-2mm}
\eqn{
p_\theta (\rvx_0) = \int p(\rvx_M) \prod_{m=1}^{M} p_\theta (\rvx_{m-1} \mid \rvx_{m}) \, \mathrm{d}\rvx_{1:M}.
}
Here, $\bx_0$ is a datapoint and $\bx_{1:M}$ are latent variables of the same dimensionality as $\bx_0$. A diffusion model consists of two core processes: the reverse process and the forward process. The reverse process is defined as
\vspace{-1mm}
\eqn{
p_\ta(\bx_{m-1}|\bx_m) := \cN(\bx_{m-1}| \bmu_\ta(\bx_m,m), \sig_m\bI)\ .
}
This process transforms a noise sample $\bx_M \sim p(\bx_M) = \cN(0,\bI)$ into an observation $\bx_0$ via a sequence of denoising steps $p_\theta(\bx_{m-1}|\bx_m)$ for $m = M, \dots, 1$. In the reverse direction, the forward process defines the approximate posterior $q(\bx_{1:M}|\bx_0) = {\smash{ \prod_{m=0}^{M-1}}} q(\rvx_{m+1}|\rvx_m)$ via the forward transitions: 
\vspace{-2mm}
\eqn{
q(\rvx_{m+1}|\rvx_m) := \cN(\bx_{m+1}; \sqrt{\al_m}\bx_m, (1-\al_m)\bI)\ .
}
The forward process iteratively applies this transition from $m=0,...,M\!-\!1$ according to a predefined variance schedule $\alpha_1, \dots, \alpha_M$ and gradually transforms the observation $\bx_0$ into noise $\cN(0,\bI)$ as $m \rightarrow M$ for a sufficiently large $M$. Unlike the reverse process involving learnable model parameters $\ta$, the forward process is predefined.
Learning the parameter $\theta$ of the reverse process is done by optimizing the variational lower bound on the log likelihood $\log p_\theta(\bx_0)$. \citet{ho2020denoising} demonstrated that this can be achieved by minimizing the following simple denoising objective:
$\Ls(\theta) = \E_{\rvx_0, m, \bep} \left[\lVert \bep - \bep_\theta(\bx_m, m) \rVert^2 \right]$.

\textbf{Planning with Diffusion.}
A prominent approach to planning via diffusion has recently gained increasing attention, with Diffuser~\citep{janner2022diffuser} being one of the most well-known methods. Diffuser formulates planning as a trajectory generation problem using diffusion models, first training a diffusion model $p_\ta(\btau)$ on offline trajectory data $\btau = (s_0, a_0, \dots, s_T, a_T)$ and then guiding sampling towrad high-return trajectories using a classifier guided approach \citep{dhariwal2021diffusion}. This guidance model, $p_\phi(\by|\btau) \propto \exp(G_\phi(\bx))$, predicts trajectory returns, modifying the sampling distribution to  $\tp_\ta(\btau) \propto p_\ta(\btau)\exp(\cJ_\phi(\btau))$, which biases the denoising process toward optimal trajectories at test time. 

\section{Proposed Method}

Our goal is to enable planning far beyond the horizon lengths available in the original dataset. To this end, we propose a two-phase approach. First, we extend the collected trajectories to significantly longer horizons via a Progressive Trajectory Extension (PTE) procedure. Second, we train a hierarchical diffusion-based planner on these extended trajectories, which we call the Hierarchical Multiscale Diffuser (HM-Diffuser). This planner is specifically designed to handle multiple temporal scales efficiently, significantly improving long-horizon planning capabilities.

\subsection{Progressive Trajectory Extension}\label{sec:PTE}

\begin{figure*}[t]
    \centering
    \includegraphics[width=0.9\textwidth]{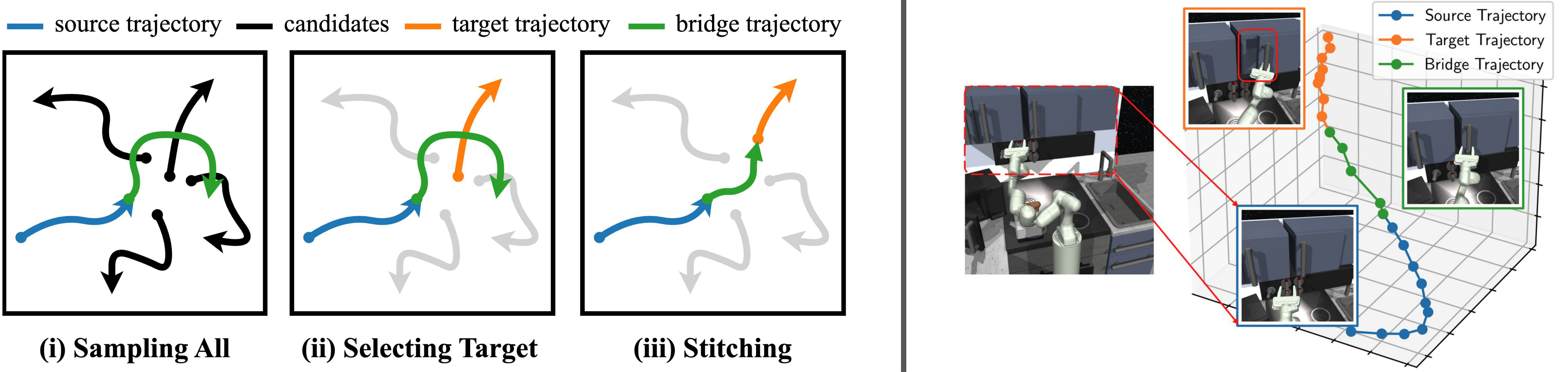}
    \vspace{-0.3cm}
    \caption{
        \textbf{Progressive Trajectory Extension process.} 
        \textbf{(Left)} Conceptual illustration of PTE: (i) A source trajectory (blue), a bridge (green), and target candidates (black) are sampled. (ii) Candidates are filtered by proximity to the bridge, and a valid target (orange) is selected. (iii) The extended trajectory is formed by bridging the source to the chosen target.
        \textbf{(Right)} Visualization of this process in the Franka Kitchen environment.
    }
    \vspace{-.5cm}
    \label{fig:stitching_diagram}
\end{figure*}

\textbf{Progressive Trajectory Extension (PTE)} is an iterative process that stitches together shorter trajectories into longer ones through successive rounds. Before running PTE, we train a few key components needed for trajectory stitching: (1) a diffusion model  called the "stitcher" for bridging trajectories, (2) an inverse dynamics model for inferring actions, and (3) a reward model for estimating rewards. The stitcher $p_\ta^\text{stitcher}(\btau)$ is trained on the base dataset $D_0$, composed of short trajectories collected from the environment, to generate trajectory segments from a given start state. The inverse dynamics model $f_a$ predicts an action $a_t$ from consecutive states $(s_t, s_{t+1})$, and the reward model $f_r$ estimates the reward $r_t$ for a state-action pair $(s_t, a_t)$. These components lay the groundwork for extending trajectories.

For each round of PTE, the algorithm takes as input a source set $\cS$ of trajectories to extend and a target set $\cT$ of trajectories providing potential continuations. How $\cS$ and $\cT$ are constructed plays a crucial role in determining the extension behavior of the resulting trajectories. In Section~\ref{sec:tech}, we describe several effective strategies for organizing these sets to guide trajectory growth. As illustrated in Figure~\ref{fig:stitching_diagram}, each PTE round performs a series of stitching iterations as follows:

\textbf{(i) Sampling a source trajectory, a bridge, and target candidates.}
We begin by randomly selecting a \textbf{source} trajectory $\btau_s \in \cS$ for extension. Next, we generate a \textbf{bridge} by conditioning the stitcher model on the last state of the source, $\btau_b \sim p_\ta^\text{stitcher}(\btau | s_0 = s_{\btau_s}^{\text{last}})$.
We use the reward model $f_r$ and the inverse dynamics model $f_a$ to infer rewards and actions for the bridge. Finally, we sample a batch of $c$ target \textbf{candidate} trajectories $({\btau_t^{\scriptscriptstyle(1)}}, {\btau_t^{\scriptscriptstyle(2)}}, \dots, {\btau_t^{\scriptscriptstyle(c)}}) \subset \cT$. These three components—the source, the bridge, and the candidates—form the basis for the stitching process in each round of PTE.

\textbf{(ii) Selecting a target trajectory.}
Note that not all candidate targets will be reachable by the sampled bridge. We therefore compute the minimum distance (e.g., Euclidean or cosine similarity) between the bridge trajectory $\btau_b$ and each candidate $\btau_t^{\scriptscriptstyle(i)}$. We then select one candidate whose distance to the bridge is below a threshold, for example by choosing uniformly at random among the valid ones.

\textbf{(iii) Stitching the source and the target.} 
After selecting a target based on the initial bridge, we refine the connection to improve quality. In the target trajectory $\btau_t$, there exists a state that is closest to the bridge $\btau_b$; suppose this is the $k$-th state. Then, we sample a new bridge $\btau_b^{\prime}$ from the stitcher, conditioned on both the last state of $\btau_s$ and the first $k$ states of $\btau_t$, to provide a smoother and more coherent connection. Finally, we concatenate the source, the refined bridge, and the target to form the extended trajectory: $\btau^\text{new} = \btau_s \mathbin\Vert \btau_b^{\prime} \mathbin\Vert \btau_t$. The full procedure is described in Algorithm~\ref{alg:pte}.
 

\subsubsection{Trajectory extension strategies}\label{sec:tech}
While PTE is a general framework with flexible instantiations, its behavior largely depends on how the source and target sets are constructed in each round. Since our goal is to extend the planning horizon, we design the following two strategies to progressively increase trajectory length. 

\textbf{Linear Extension} is the default strategy for steadily increasing trajectory length over successive PTE rounds. In this approach, the source set consists of trajectories extended in the previous round, while the target set remains fixed as the original base dataset: $\cS = \cD_{r-1}$, $\cT = \cD_0$. This setup ensures that each round yields slightly longer trajectories than the last, enabling stable and incremental growth.

\textbf{Exponential Extension} accelerates trajectory growth by using the output of the previous round as both the source and target sets: $\cS_r = \cT_r =  \cD_{r - 1}$. By stitching together already-extended trajectories, this strategy enables faster horizon expansion, which is beneficial in large or complex environments. However, it may result in a sparser distribution of intermediate lengths compared to the linear case. We analyze this trade-off in more detail in Section~\ref{sec:exp:maze}.

After $R$ rounds of PTE, we obtain an extended dataset $\cD_R$ with trajectories significantly longer than those in $\cD_0$. This dataset is then used to train our hierarchical planner, designed for effective long-horizon planning. 

\begin{figure*}[t]
    \centering
    \includegraphics[width=0.95\textwidth]{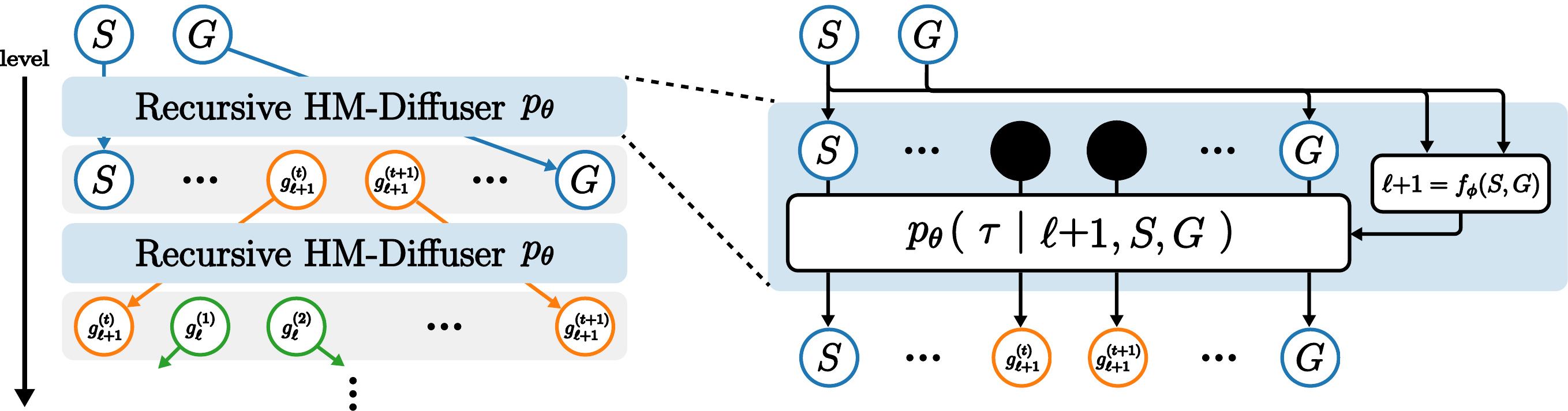}
    \caption{\textbf{Recursive Hierarchical Multiscale Diffuser (HM-Diffuser).} Given a start state S and goal state G, the Adaptive Plan Pondering (APP) module selects an appropriate starting level $\ell\text{+}1 = f_\phi(S, G)$. The level-conditioned diffuser $p_\ta( \tau | \ell\text{+}1, S, G)$ generates a sequence of high-level subgoals, which are recursively refined by the same model at lower levels. Each subgoal pair defines a start–goal pair for the next level, progressively constructing a complete trajectory from coarse to fine resolution.}
    \vspace{-.5cm}
    \label{fig:model_archiecture}
\end{figure*}
\subsection{Hierarchical Multiscale Diffuser}
A straightforward way to train a planner on the extended dataset $D_R$ is to learn a single diffusion model to generate entire long trajectories using the standard Diffuser~\citep{janner2022diffuser,decision_diffuser}. However, scaling Diffuser to very long horizons poses serious challenges: the trajectory length directly affects the model’s output dimensionality and the complexity of denoising, leading to much higher computational cost. Moreover, prior studies~\citep{hd} has shown that Diffuser’s performance degrades as the planning horizon grows, due to the difficulty of modeling extremely long sequences.

To address these issues, we adopt the Hierarchical Diffuser (HD) framework~\citep{hd}, which improves efficiency and scalability through structured planning. Our planner consists of a hierarchy of $L$ levels, where each level $\ell \in {1, \cdots, L}$ employs a diffusion model $p_{\theta_\ell}(\btau)$. The effective planning horizon at level $\ell$ is determined by a jump length $j_\ell$ and a jump count $k_\ell$, giving $H_\ell = j_\ell \times k_\ell$. For example, if $j_\ell = 10$, meaning the planner outputs one state every 10 steps, and $k_\ell = 5$, generating five such states, the resulting trajectory spans 50 steps in total.


This temporally sparse structure enables efficient planning, as each level generates only every $j_\ell$-th state over $k_\ell$ steps, resulting in an effective horizon of $H_\ell$ with reduced output dimensionality. The resulting states, denoted as ${g_{\ell}^{\scriptscriptstyle(1)}, \dots, g_{\ell}^{\scriptscriptstyle(k_\ell)}}$, serve as \textbf{subgoals} for lower levels.


The core idea of hierarchical planning in HD is that each level $\ell$ generates a jumpy trajectory conditioned on two consecutive subgoals from the higher level $\ell+1$, with one as the start and the other as the end. Given subgoals $g_{\ell+1}^{\scriptscriptstyle(t)}$ and $g_{\ell+1}^{\scriptscriptstyle(t+1)}$, the level-$\ell$ planner generates a sequence as follows:
\[
( g_{\ell}^{\scriptscriptstyle(1)}, \dots, g_{\ell}^{\scriptscriptstyle(k_\ell-1)}, g_{\ell}^{\scriptscriptstyle(k_\ell)}) \sim p_{\ta_\ell}(\btau | g_{\ell}^{\scriptscriptstyle(1)}=g_{\ell+1}^{\scriptscriptstyle(t)}, g_{\ell}^{\scriptscriptstyle(k_{\ell})} = g_{\ell+1}^{\scriptscriptstyle(t+1)})
\]

This hierarchical structure ensures that lower-level plans are consistent with higher-level subgoals. To maintain alignment across levels, we set $H_\ell = j_{\ell+1}$, so that each jump at level $\ell+1$ is decomposed into $k_\ell$ subgoals at level $\ell$. At the lowest level, we set $j_1 = 1$, producing a fine-grained, dense plan.

\subsubsection{Adaptive Plan Pondering} 
A limitation of the above hierarchical planner is that it always starts from the top level $L$, regardless of the distance to the goal. This can lead to inefficiencies when the goal is nearby, as it produces unnecessarily indirect plans. To address this, we introduce \textbf{Adaptive Plan Pondering (APP)}, which dynamically adjusts the starting level based on the goal proximity. Specifically, we train a depth predictor $f_\phi$ that takes the start and goal states $(s_0, s_g)$ and outputs the most suitable planning level $\bar{\ell} = f_\phi(s_0, s_g)$. Since the target level for each trajectory in the training dataset $\cD_R$ can be readily determined (e.g., from its length), the predictor can be trained in a straightforward supervised manner.

At test time, APP enables the planner to skip unnecessary higher levels and begin planning from a lower level when appropriate. This not only reduces computational overhead but also improves overall plan quality by avoiding unnecessary detours.


\subsubsection{Recursive HM-Diffuser} 
\label{subsec:recursive_HMD}
Another key limitation of the hierarchical approach is the need to maintain separate diffuser models for each level—$p_{\ta_1}, p_{\ta_2}, \dots, p_{\ta_L}$—each with its own set of parameters. This raises an important question: can a single diffusion model effectively handle all levels of the hierarchy? While sharing parameters may not necessarily improve performance compared to non-shared models with larger capacity, it significantly simplifies model management and reduces computational complexity, making it a desirable approach if performance can be maintained.

To that end, we incorporate Recursiveness into HM-Diffuser, enabling a single diffusion model to handle the entire hierarchy, as illustrated in Figure~\ref{fig:model_archiecture}. Instead of maintaining separate diffusers for each level, we replace them with a single level-conditioned diffusion model $\smash{p_\ta(\btau|\ell)}$. During training, the shared diffuser is trained to generalize across different levels by uniformly sampling $\smash{\ell \sim \text{Uniform}(1,\dots,L)}$, ensuring exposure to all levels and enabling it to generate trajectories conditioned on any level.

For planning, the process starts at the appropriate level predicted by APP. Once a high-level subgoal sequence is generated, each pair of consecutive subgoals $(g_{\ell+1}^{\scriptscriptstyle(t)}, g_{\ell+1}^{\scriptscriptstyle(t+1)})$ is recursively fed back into the same diffuser, with the level indicator explicitly decreased by one from $\ell+1$ to $\ell$. The model then generates a refined trajectory at the lower level, conditioned on a start and end state defined by a pair of consecutive states from the higher level:
$\smash{p_\ta(\btau | \ell, g_{\ell}^{\scriptscriptstyle(1)}=g_{\ell+1}^{\scriptscriptstyle(t)}, g_{\ell}^{\scriptscriptstyle(k_\ell)} = g_{\ell+1}^{\scriptscriptstyle(t+1)})}$.
This process continues until the lowest level is reached, resulting in a fully specified dense trajectory.

\section{Related Works}

\textbf{Diffusion-based planners in offline RL.} Diffusion models are powerful generative models that frame data generation as an iterative denoising process \citep{ho2020denoising,ddim}. They were first introduced in reinforcement learning as planners by \cite{janner2022diffuser}, utilizing their sequence modeling capabilities. Subsequent work \citep{decision_diffuser, liang2023adaptdiffuser, rigter2023world} has shown promising results in offline-RL tasks. Diffusion models have also been explored as policy networks to model multi-modal behavior policies \citep{wang2023diffusion, kang2024efficient}. Recent advancements have extended these models to hierarchical architectures \citep{wenhao2023hdoffline, hd, dong2024diffuserlite, chen2024plandq}, proving effective for long-horizon planning. Our method builds on this by not only using diffusion models for extremely long planning horizons but also exploring the stitching of very short trajectories with them. Additional related works on hierarchical planning are provided in Appendix~\ref{apdx:related}.

\textbf{Data augmentation in RL} has been a crucial strategy for improving generalization in offline RL. Previous work used dynamic models to stitch nearby states \citep{char2022bats}, generate new transitions \citep{hepburn2022modelbased}, or create trajectories from sampled initial states \citep{zhou2023free, lyu2022double, wang2021offline,zhang2023uncertainty}. Recently, diffusion models have been applied for augmentation \citep{zhu2023diffusion}. \cite{lu2023synthetic} used diffusion models to capture the joint distribution of transition tuples, while \cite{he2024diffusion} extended it to multi-task settings. \cite{diffstitch} used diffusion to connect trajectories with inpainting.

\section{Experiments}\label{sec:exp}
\begin{figure*}[t]
    \centering
    \begin{minipage}{0.26\textwidth}  
        \centering
        \includegraphics[height=0.14\textheight]{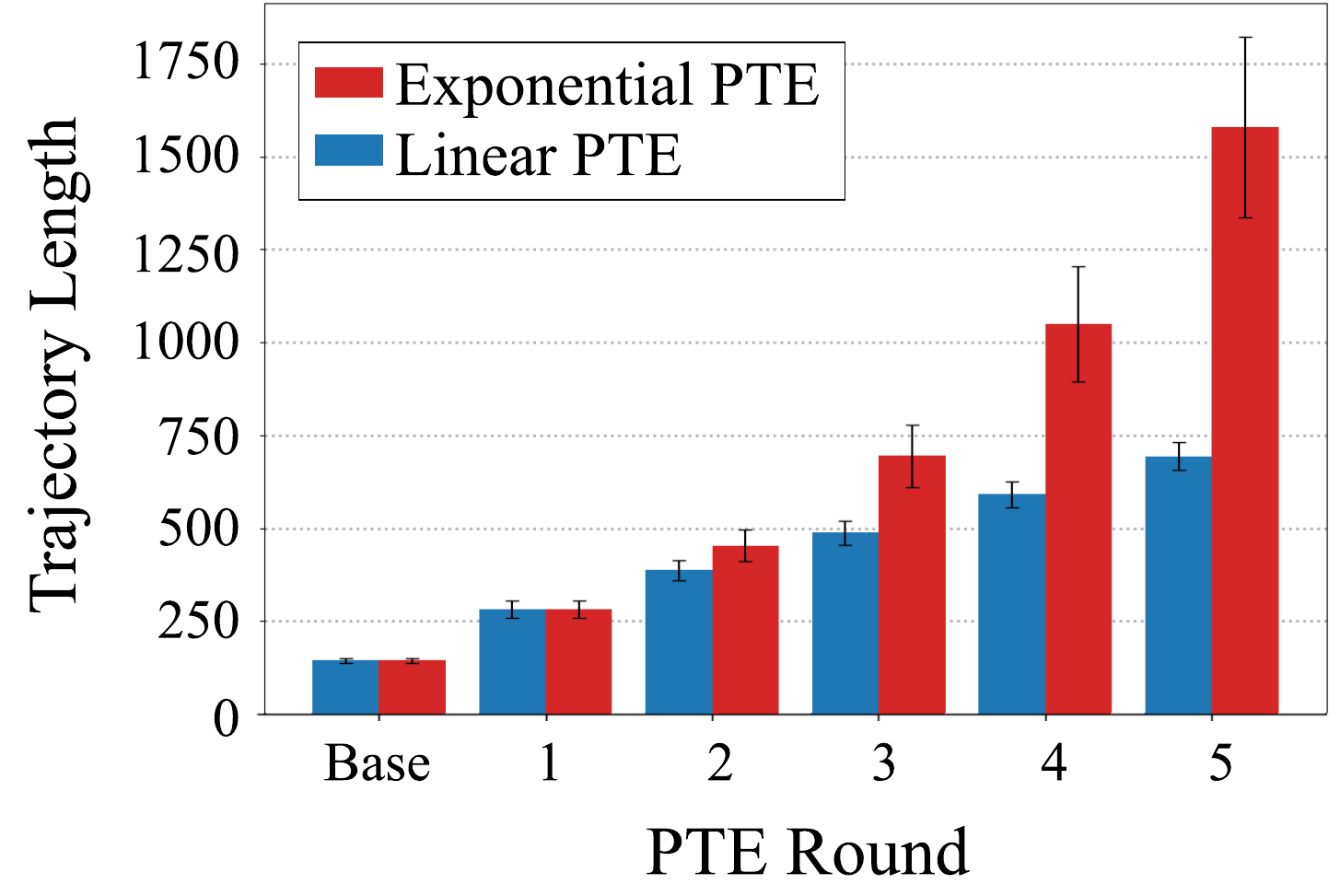}
        \caption*{\hspace{7mm} (a) Mean Trajectory Length}  
        \label{fig:mean_traj_length}  
    \end{minipage}  
    \hspace{5mm}
    \begin{minipage}{0.26\textwidth}  
        \centering
        \includegraphics[height=0.14\textheight]{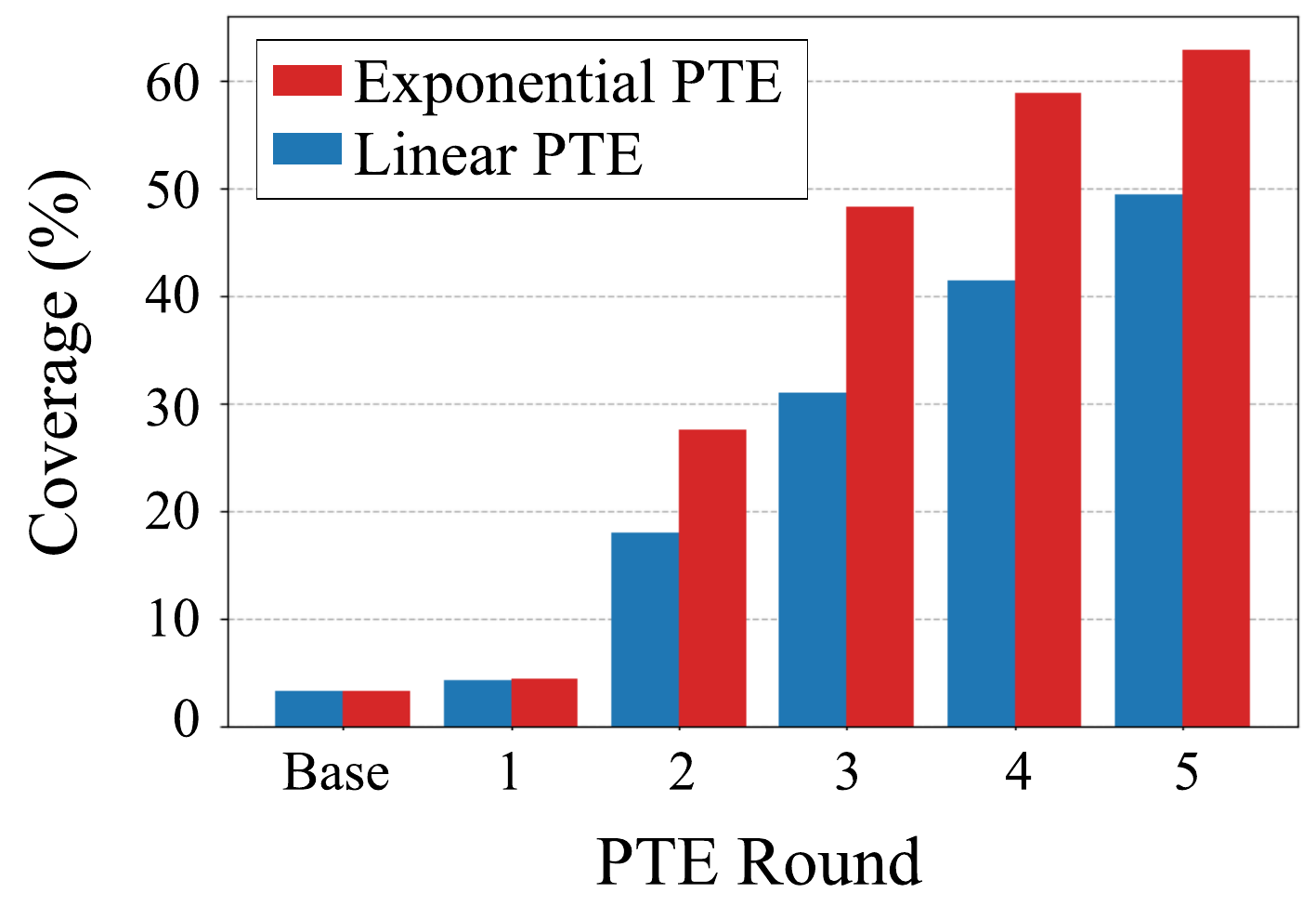}
        \caption*{\hspace{8mm} (b) Start-Goal Pair Coverage}  
        \label{fig:path_coverage}
    \end{minipage}    
    \begin{minipage}{0.42\textwidth}  
        \centering
        \includegraphics[height=0.14\textheight]{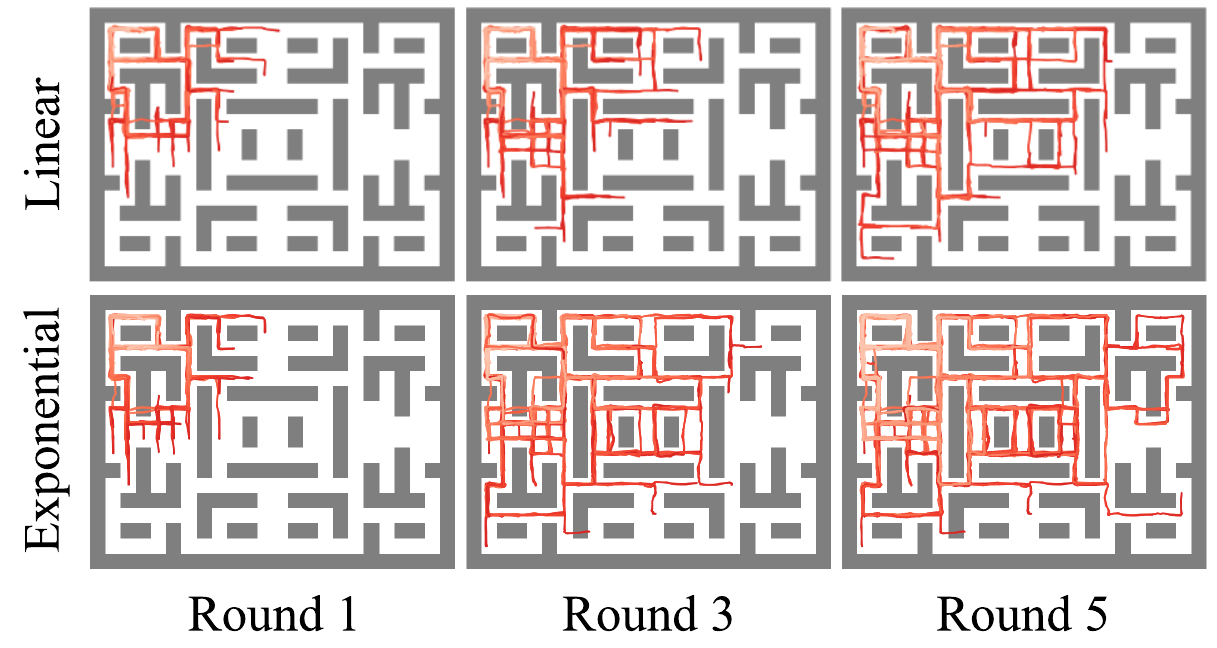}
        \caption*{\hspace{1mm} (c) Maze Coverage over PTE Rounds}  
        \label{fig:qualitative_maze_coverage}
    \end{minipage}
    \vspace{-2mm}

    \caption{\textbf{Progressive Trajectory Extension results.} (a) Mean trajectory lengths for the XXLarge maze. Linear PTE increases trajectory length at a constant rate, while Exponential PTE grows exponentially across rounds. (b) Start-goal pair coverage improves as PTE progresses, with Exponential PTE covering a significantly larger portion of the maze compared to Linear PTE. (c) Maze coverage visualizations over PTE rounds. Exponential PTE leads to faster and broader coverage than Linear PTE.}
    \label{fig:stitching_quantitative}
    \vspace{-2mm}
\end{figure*}



We aim to investigate the following questions through our experiments:
(1) Can PTE generate trajectories that are significantly longer than those originally collected from the environment? 
(2) Can HM-Diffuser leverage these extended trajectories to produce coherent long-horizon plans? 
(3) Is the overall approach effective in high-dimensional manipulation tasks? 
To facilitate our analysis, we introduce the Plan Extendable Trajectory Suite (PETS).

\subsection{Plan Extendable Trajectory Suite}\label{sec:pets}

In practice, collecting long trajectories is often costly and impractical, whereas short-horizon data is far more accessible. However, existing benchmarks rarely evaluate the ability to perform long-horizon planning when only short-horizon data is available. To address this gap, we introduce the \textbf{Plan Extendable Trajectory Suite (PETS)}—a benchmark built by restructuring existing environments to evaluate this capability. PETS spans three domains—Maze2D, Franka Kitchen, and Gym-MuJoCo—each chosen to represent a distinct challenge in control and generalization.



\textbf{Extendable Maze2D.}
Maze2D is a widely used environment for evaluating planning algorithms. Building on the existing Large Maze from D4RL and the Giant Maze from \citet{park2024ogbench}, we additionally introduce a new XXLarge Maze, four times larger than the Large Maze, to enable more rigorous evaluation of long-horizon planning. To assess extendable planning capability, training data consists only of short trajectories between 130 and 160 steps. However, at test time, tasks require reaching goals up to 1,000 steps, depending on the maze size. See Appendix~\ref{app:details_maze2d} for details.

\textbf{Extendable Franka Kitchen.} FrankaKitchen is a widely used benchmark for complex, high-dimensional manipulation tasks. We modify the training data by splitting the D4RL offline trajectories into non-overlapping segments of length 10, exposing the model only to short-horizon trajectories during training. At test time, however, it must complete compound tasks that require much longer sequences involving multiple stages of manipulation. This setup enables evaluation of extendable long-horizon planning while preserving the full complexity of the original high-dimensional environment. See Appendix~\ref{app:additional_kitchen_dmc} for details.

\textbf{Extendable Gym-MuJoCo.}
Gym-MuJoCo environments are widely used benchmarks for low-level motor control, but they are not originally designed for long-horizon planning. To adapt them, we split the D4RL offline trajectories from Hopper and Walker2D into non-overlapping segments of length 10, so that only short-horizon trajectories are available during training. At test time, however, models are required to plan over much longer horizons while maintaining temporally coherent control. This setup retains the dense reward structure of the original environments while enabling assessment of extendable long-horizon planning. See Appendix~\ref{app:additional_kitchen_dmc} for details.

We now present experimental results based on the PETS benchmark. Throughout this section, we use the -X suffix to denote baselines trained on the extended dataset generated by PTE, which reflects their ability to model long-horizon plans. In contrast, baselines without this suffix are trained solely on the short base dataset. Additional implementation details are provided in Appendix~\ref{app:implementation_details}.


\subsection{Extendable Maze2D results}
\label{sec:exp:maze}
We first validate our proposed unified framework of PTE and HM-Diffuser in the Extendable Maze2D environment, focusing on whether PTE can generate trajectories that connect unseen start-goal pairs to improve maze coverage, and whether HM-Diffuser can utilize them to reach goals situated in remote areas of the maze.

\textbf{PTE on Extendable Maze2D.} As shown in Figure~\ref{fig:stitching_quantitative}(a), linear and exponential variants of PTE progressively and substantially increase trajectory lengths over successive rounds. While both methods show similar gains in the early stages, their growth dynamics differ: Linear PTE increases length at a steady rate, reaching roughly 700 steps by round 5, whereas exponential PTE accelerates rapidly, exceeding 1,500 steps at the same round as trajectories from previous rounds are stitched together to form even longer ones. 

This trajectory extension leads to broader maze coverage, as illustrated in Figure~\ref{fig:stitching_quantitative}(b). Both variants significantly expand the diversity of reachable start-goal pairs compared to the base dataset, which covers less than 5\%. By round 5, linear PTE increases coverage to over 65\%, and exponential PTE further extends it beyond 80\%. However, as detailed in Section~\ref{exp:ablation}, since exponential PTE requires more data than Linear one, we adopt linear PTE as the default setting in our experiments due to its stability and data efficiency. In the following experiments, we apply 7 rounds of Linear PTE to extend trajectory lengths. The resulting datasets are then aggregated into a single training set.
\begin{table}[t]
  \centering
  \caption{\textbf{Extendable Maze2D performance.} We compare the performance of baselines in Extendable Maze2D and Extendable Multi2D settings in various maze environments (Large, Giant, and XXLarge). The -X baselines are trained on the extended dataset generated by applying 7 rounds of Linear PTE, while Diffuser is trained only on the short base dataset. We report the mean and the standard error over 200 planning seeds. HM-Diffuser-X consistently outperforms all baselines across all settings.}
  \begin{adjustbox}{width=1\linewidth}
    \begin{tabular}{llrrrr}
    \toprule 
    
     \multicolumn{2}{c}{\textbf{Environment}}   & \textbf{Diffuser} &\textbf{Diffuser-X}      & \textbf{HD-X}         & \textbf{HMD-X}    \\ \midrule
    
    Maze2D                  & Large & $45.8 \pm 5.9$ & $114.4 \pm 4.7$ & $144.3 \pm 3.3$ & $\mathbf{166.9} \pm 4.6$                \\
    Maze2D                  & Giant  & $77.2 \pm 11.7$ & $114.6 \pm 9.10$ & $138.4 \pm 9.6$ & $\mathbf{177.4} \pm 11.9$                \\
    Maze2D                  & XXLarge & $14.1 \pm 4.9$ & $23.0 \pm 3.4$ & $56.4 \pm 4.9$ & $\mathbf{82.1} \pm 8.5$               \\ \midrule
    \multicolumn{2}{c}{\textbf{Single-task Average}}    & $45.7$ & $84.0$ & $113.0$ & $\mathbf{142.1}$ \\ \midrule
    Multi2D             & Large              & $33.4 \pm 5.9$ & $130.3 \pm 4.0$ & $150.3 \pm 3.0$ & $\mathbf{174.7} \pm 3.8$     \\ 
    Multi2D             & Giant              & $77.8 \pm 11.8$ & $140.2 \pm 8.9$ & $168.8 \pm 9.0$ & $\mathbf{246.7} \pm 10.6$    \\ 
    Multi2D             & XXLarge              & $23.4 \pm 6.3$ & $63.2 \pm 5.2$ & $71.9 \pm 5.5$ & $\mathbf{109.9} \pm 8.4$      \\ \midrule
    \multicolumn{2}{c}{\textbf{Multi-task Average}}  & $44.9$ & $111.2$ & $130.4$ & $\mathbf{177.1}$
    \\
    \bottomrule
    \end{tabular}
    \end{adjustbox}
  \label{table:maze2d_performance}
  \vspace{-0.7cm}
\end{table}

\begin{figure*}[t!]
    \centering
    \includegraphics[width=1.0\textwidth]{images/inkscape/neurips_hmd_comparison.pdf}
    \vspace{-3mm}

    \caption{\textbf{Adaptive planning capability of HM-Diffuser.} (a) Multiscale planning trajectories generated by HM-Diffuser across different levels ($\ell=1,2,3$). The orange box highlights the level selected by the depth predictor, which yields the most efficient plan. (b) Comparison of long-horizon planning results from Diffuser, HD, and HM-Diffuser. While Diffuser and HD exhibit detours or suboptimal paths, HM-Diffuser generates a more direct and optimal trajectory.}
    \label{fig:trajectory_planning_vis}
    \vspace{-5mm}
\end{figure*}

\textbf{HM-Diffuser on Extendable Maze2D.} Using the extended dataset, we train Diffuser, HD, and our proposed HM-Diffuser, denoting these models with the -X suffix. Following Diffuser~\citep{janner2022diffuser}, we evaluate performance under two settings. In the single-task setting (Extendable Maze2D), the goal position is fixed at the bottom-right corner of the maze while the start position is randomized. The multi-task setting (Extendable Multi2D) randomizes both start and goal positions to assess the model’s generalization across a wider range of tasks. Evaluation details are provided in Appendix~\ref{app:details_maze2d}.

Table \ref{table:maze2d_performance} demonstrates that HM-Diffuser-X consistently outperforms all baselines across both settings. In the single-task setting, HM-Diffuser-X achieves the highest average score ($142.1$), highlighting the effectiveness of its hierarchical and multiscale planning capabilities. Meanwhile, Diffuser-X, despite being trained on the extended dataset, struggles in the XXLarge maze environment, achieving only $23.0 \pm 3.4$—a performance comparable to Diffuser trained solely on the base dataset. This suggests that Diffuser-X faces challenges in modeling long-horizon plans in complex environments. However, HD-X and HM-Diffuser-X manage to retain better performance, demonstrating the effectiveness of the hierarchical planning. In the multi-task setting, HM-Diffuser-X outperforms all baselines, achieving an average score of $177.1$, demonstrating its ability to generalize across diverse tasks.

These results demonstrate that PTE effectively constructs longer and more diverse trajectories, enabling HM-Diffuser to plan over substantially longer horizons than those present in the original dataset. The combination of scalable data generation via PTE and efficient hierarchical planning via HM-Diffuser proves essential for solving complex, long-horizon tasks. Detailed results across different PTE rounds are presented in Figure~\ref{fig:app:performance_rounds_maze2d_full}.


\subsection{Extendable Franka Kitchen and Extendable Gym-MuJoCo results}
We also evaluate our framework on Extendable Franka Kitchen and Extendable Gym-MuJoCo to assess its ability to generalize beyond the short training trajectories and produce coherent plans over extended horizons in both high-dimensional manipulation and low-level locomotion control.

\textbf{PTE on Extendable Franka Kitchen and Extendable Gym-MuJoCo.}
To assess the effect of PTE in these complex environments, we analyze the distribution of normalized returns—defined as total return divided by trajectory length. This metric provides a conservative estimate of data quality: if longer trajectories are generated without meaningful improvement, the normalized return would decline. As shown in Figure~\ref{fig:dmc_return_distribution}, PTE shifts the distribution upward, increasing the density of high-return trajectories and indicating improved overall data quality. Despite a slight dip in Round 1 for Extendable Hopper-Medium-Replay, likely attributed to dataset characteristics, the normalized return improves in Round 2, demonstrating the effectiveness of PTE over time.

\textbf{HM-Diffuser on Extendable Franka Kitchen.}
As shown in Table~\ref{exp:performance_kitchen_dmc}, training on the extended dataset improves the performance of Diffuser-X from $43.8$ to $45.8$, likely due to the longer planning horizon enabled by PTE, which addresses short-horizon limitations by connecting short trajectories into longer ones that complete the task (see Appendix~\ref{app:fig:kitchen_stitch}). Both HM-Diffuser-X and HD-X outperform Diffuser-X by leveraging hierarchical planning that reduces complexity—a trend also observed in \citet{chen2024simple}. Between the two, HM-Diffuser-X scores higher than HD-X: $59.2$ vs.\ $55.0$, demonstrating the benefit of our efficient parameter sharing and adaptive planning design.

\begin{table}[t!]
  \begingroup
  \setlength{\tabcolsep}{4pt} 
  \renewcommand{\arraystretch}{1.2}
  \centering
  \caption{\textbf{Performance on Extendable Franka Kitchen and Extendable Gym-MuJoCo.} Training on the extended dataset generally improves the average performance of Diffuser-X. Introducing a recursive hierarchical structure further enhances the performance. Results are averaged over 15 planning seeds.}
  \vspace{-0.1cm}
  \begin{adjustbox}{width=1\linewidth}
    \begin{tabular}{llrrrr}
    \toprule 
    \multicolumn{2}{c}{\textbf{Environment}}        & \textbf{Diffuser}    &\textbf{Diffuser-X}    & \textbf{HD-X}     & \textbf{HMD-X}    \\ \midrule
    Kitchen     & Partial-v0    & $41.7 \pm 3.2$        & $43.3 \pm 5.5$        & $\mathbf{56.7} \pm 5.8$    & $\mathbf{56.7} \pm 5.3$                \\
    Kitchen     & Mixed-v0      & $45.8 \pm 3.1$        & $48.3 \pm 4.7$        & $53.3 \pm 3.1$               & $\mathbf{61.7} \pm 3.1$               \\  \midrule
    \multicolumn{2}{c}{\textbf{Kitchen Average}}    & $43.8$ & $45.8$ & $55.0$ & $\mathbf{59.2}$ \\ \midrule
    Walker2d     & MedReplay    & $22.8 \pm 2.7$        & $20.1 \pm 4.3$        & $\mathbf{30.2} \pm 5.9$    & $29.6 \pm 4.8$                \\
    Walker2d     & Medium       & $58.1 \pm 5.6$        & $62.6 \pm 6.4$        & $66.5 \pm 4.3$    & $\mathbf{72.7} \pm 2.5$               \\ 
    Walker2d     & MedExpert    & $\mathbf{82.3} \pm 4.6$        & $80.3 \pm 3.7$        & ${80.8} \pm 2.9$    & $79.3 \pm 2.3$               \\ \midrule
    \multicolumn{2}{c}{\textbf{Walker2d Average}}    & $54.4$ & $54.3$ & $59.2$ & $\mathbf{60.5}$ \\ \midrule
    Hopper      & MedReplay    & $18.7 \pm 3.0$       & $34.5 \pm 6.2$        & $22.5 \pm 3.1$    & $\mathbf{37.3} \pm 4.8$                \\
    Hopper      & Medium       & $\mathbf{45.6} \pm 1.9$        & $44.3 \pm 3.5$        & $44.1 \pm 2.8$    & ${44.9} \pm 3.5$               \\ 
    Hopper      & MedExpert    & $61.4 \pm 8.4$        & $\mathbf{74.9} \pm 8.0$        & $67.9 \pm 7.7$    & $74.3 \pm 9.0$               \\ \midrule
    \multicolumn{2}{c}{\textbf{Walker2d Average}}    & $41.9$ & $51.2$ & $44.8$ & $\mathbf{52.2}$ \\ 
    \bottomrule
    \end{tabular}
    \end{adjustbox}
  \label{exp:performance_kitchen_dmc}
  \endgroup
  \vspace{-0.3cm}
\end{table}
\begin{figure}
    \hspace{-4mm}
    \centering
    \includegraphics[width=0.49\textwidth]{images/inkscape/neurips_GymMujoco_Vis_no_version_extendable2.pdf}
    \vspace{-6mm}
    \caption{\textbf{Trajectory quality.} To fairly compare trajectory quality as length increases with stitching, we use normalized return (total return divided by its length). Although this metric tends to penalize longer trajectories, it still improves across PTE rounds, indicating that PTE progressively generates higher-quality data.}
    \label{fig:dmc_return_distribution}
    \vspace{-.5cm}
\end{figure}
\textbf{HM-Diffuser on Extendable Gym-MuJoCo.}
As shown in Table~\ref{exp:performance_kitchen_dmc}, training on the extended dataset improves the average performance of Diffuser-X on Extendable Hopper from $41.9$ to $51.2$, likely due to PTE alleviating limitations from short-horizon training data. On Extendable Walker2D, Diffuser-X performs similarly to Diffuser ($54.3$ vs. $54.4$), showing no notable gain despite access to extended trajectories. However, prior work \citep{chen2024simple} reports that standard Diffuser models often suffer performance degradation when planning over long horizons due to increased complexity. In contrast, our results show that Diffuser-X maintains stable performance, suggesting that PTE mitigates such degradation by generating higher-quality training data. Notably, HM-Diffuser-X outperforms both Diffuser-X and HD-X, achieving average scores of $60.5$ on Extendable Walker2D and $52.2$ on Extendable Hopper.


\subsection{Ablation studies}
\label{exp:ablation}
To better understand the effectiveness of PTE and HM-Diffuser, we conduct ablations on key design choices, including the adaptive planning mechanism, the impact of different extension strategies, and comparisons with standard benchmarks and alternative data augmentation methods.


\textbf{Linear PTE and Exponential PTE.} We compare Linear and Exponential PTE on the Extendable Maze2d-XXLarge task and find that Exponential PTE is highly sensitive to dataset size. Linear PTE shows steady performance improvements as the dataset size grows from 1M to 4M, with scores rising from $82.1\pm8.5$ to $89.9\pm8.2$ using data generated from 7 PTE rounds. In contrast, Exponential PTE improves dramatically from $30.0 \pm 5.5$ to $120.5 \pm 8.6$ over the same range, highlighting its dependence on larger datasets. As our goal is to evaluate general applicability rather than maximize absolute performance, we adopt Linear PTE as the default. Detailed results are provided in Table~\ref{app:exp:maze2d_performance_exponential_datasetSize}.

\textbf{PTE outperforms other diffusion-based data augmentation methods.} Next, we compare our proposed PTE with DiffStitch \citep{diffstitch}, a diffusion-based data augmentation method, on the extendable planning problem. Table~\ref{app:exp:pte_diffsttich} shows that all planners trained in PTE-augmented datasets outperform those trained in DiffStitch-augmented datasets. For further analysis, please see Appendix~\ref{app:sec:pte_vs_diffstitch}.

\textbf{Adaptive Plan Pondering.} In Figure \ref{fig:trajectory_planning_vis} (a), we qualitatively illustrate the multiscale planning capability of HM-Diffuser, conditioned on different levels, where $\ell = 1$ represents the lowest level. With the appropriate level determined by the pondering depth predictor, HM-Diffuser maintains effective planning without deterioration, as indicated by the orange box in Figure \ref{fig:trajectory_planning_vis} (a) (see Figure \ref{fig:app:multiscale_planning} for additional examples). Additionally, Figure \ref{fig:trajectory_planning_vis} (b) shows that Diffuser and HD generate inefficient plans with excessive detours due to their fixed planning horizon (refer to Figure \ref{fig:app:trajectory_planning_methods} for more examples). Unlike other methods, HM-Diffuser performs adaptive planning by selecting the appropriate level per task. Figure~\ref{fig:app:hier} illustrates how hierarchical planning proceeds from the predicted level down to lower levels, enabling efficient and flexible trajectory generation.


\textbf{HM-Diffuser performs reasonably well on standard benchmarks.} Finally, we evaluate the proposed HM-Diffuser on the standard D4RL benchmark. Unlike previous experiments using PTE-processed data, the trajectories here are taken directly from the original D4RL dataset without splitting. Table \ref{app:performance_standard_benchmark} demonstrates that HM-Diffuser outperforms all baselines in the long-horizon planning tasks and performs reasonably well in standard control tasks.

\section{Conclusion}\label{sec:conclusion_limitation}
In this work, we introduce the Hierarchical Multiscale Diffuser framework for the proposed extendable long-horizon planning problem. Starting from a set of short, suboptimal trajectories, our approach employs Progressive Trajectory Extension to generate longer, more informative sequences. We then train an HM-Diffuser planner on the augmented dataset, leveraging a hierarchical multiscale structure to efficiently model and execute long-horizon plans. Experiments demonstrate that our framework achieves promising results across a diverse set of benchmarks, including the long-horizon Extendable Maze2D, dense-reward Extendable Gym-MuJoCo, and high-dimensional Extendable FrankaKitchen manipulation tasks.

While HM-Diffuser significantly improves long-horizon planning, several limitations remain. First, although our method achieves strong overall performance, further improving stitching quality could lead to even greater gains.
Second, HM-Diffuser currently operates in state space and does not support visual inputs, extending it to image-based planning is essential for real-world applications. Third, plan pondering is restricted to discrete levels, and the lack of temporal abstraction may limit scalability to extremely long horizons. Finally, HM-Diffuser lacks test-time adaptivity; integrating search-based refinement like MCTS could enhance flexibility during execution.

\section{Acknowledgment}
This research was supported by Brain Pool Plus Program (No. 2021H1D3A2A03103645) and GRDC (Global Research Development Center) Cooperative Hub Program (RS-2024-00436165) through the National Research Foundation of Korea (NRF) funded by the Ministry of Science and ICT (MSIT). This work was also partly supported by Electronics and Telecommunications Research Institute (ETRI) grant funded by the Korean government (25ZR1100) and Samsung Advanced Institute of Technology.


\bibliography{refs, ref_cc, refs_cg, refs_ahn_local}


\clearpage


\newpage
\appendix
\setcounter{secnumdepth}{2}
\renewcommand\thefigure{\thesection.\arabic{figure}}
\renewcommand\thetable{\thesection.\arabic{table}}   
\renewcommand\thealgorithm{\thesection.\arabic{algorithm}}

\onecolumn




\section{Impact Statement}\label{app:sec:impact}
This work introduces Hierarchical Multiscale Diffuser (HM-Diffuser), a framework that extends diffusion-based planning to significantly longer horizons through hierarchical multiscale modeling and Progressive Trajectory Extension (PTE). As a general-purpose planning framework, HM-Diffuser enables the creation of longer trajectories from insufficient training datasets, allowing planners to solve complex offline reinforcement learning tasks. While the framework itself does not inherently pose direct risks, its application to safety-critical and high-stakes domains, such as autonomous systems, robotics, healthcare, finance, and logistics, warrants careful ethical consideration. Long-horizon decision-making in these areas can have profound impacts, necessitating safeguards to ensure reliability, fairness, and alignment with societal values. Moreover, the increased computational demands of diffusion-based planning raise concerns about energy efficiency and sustainability, emphasizing the importance of responsible AI deployment. As this line of research advances, it is also essential to address broader ethical issues, including potential biases in decision-making, data privacy, and job displacement risks due to automation. Ensuring these technologies benefit society as a whole requires collaborative efforts among researchers, policymakers, and industry stakeholders to promote transparency, inclusivity, and equitable outcomes.

\section{Additional Related Works}
\label{apdx:related}
\textbf{Hierarchical Planning.} Hierarchical frameworks are widely used in reinforcement learning (RL) to tackle long-horizon tasks with sparse rewards. Two main approaches exist: sequential and parallel planning. Sequential methods use temporal generative models, or world models \citep{worldmodels,hafner2019learning}, to forecast future states based on past data \citep{li2022hierarchical,director,hu2023planning,zhu2023making}. Parallel planning, driven by diffusion probabilistic models \citep{janner2022diffuser,decision_diffuser}, predicts all future states at once, reducing compounding errors. This has combined with hierarchical structures, creating efficient planners that train subgoal setters and achievers \citep{hdmi,simple,dong2024diffuserlite,chen2024plandq}. 

\section{Implementation Details}\label{app:implementation_details}
In this section, we describe the architecture and the hyperparameters used for our experiments. We used a single NVIDIA RTX A5000 GPU for each experiment, with each run taking approximately 24 hours.


\begin{itemize}
    \item We build our code based on the official code release of Diffuser \cite{janner2022diffuser} obtained from \url{https://github.com/jannerm/diffuser} and official code release of HD \citep{chen2024simple} obtained from \url{https://github.com/changchencc/Simple-Hierarchical-Planning-with-Diffusion}.
    
    \item We represent the level embeddings with a 2-layered MLP with a one-hot level encoding input. We condition the diffuser on the level embedding to generate multiscale trajectories. For training, we sample different levels and the level determines the resolution of the sampled trajectories.
    \item For the stitcher model, we train the diffuser with a short horizon $H$ (\texttt{Maze2D-Large:} $80$, \texttt{Maze2D-Giant:} $80$, \texttt{Maze2D-XXLarge:} $80$, \texttt{Gym-MuJoCo:} $10$, \texttt{Kitchen:} $10$)
    \item We represent the pondering depth predictor $f_\phi^{L}(l\vert s_1, s_2)$ with a 3-layered MLP with 256 hidden units and ReLU activations. The classifier trained with samples from multiscale trajectories to predict the corresponding level.

\end{itemize}


\section{Details for Extendable Maze2D tasks}\label{app:details_maze2d}

\begin{figure*}[ht!]
    \centering
    \begin{minipage}{0.3\textwidth}
        \centering
        \includegraphics[width=\textwidth]{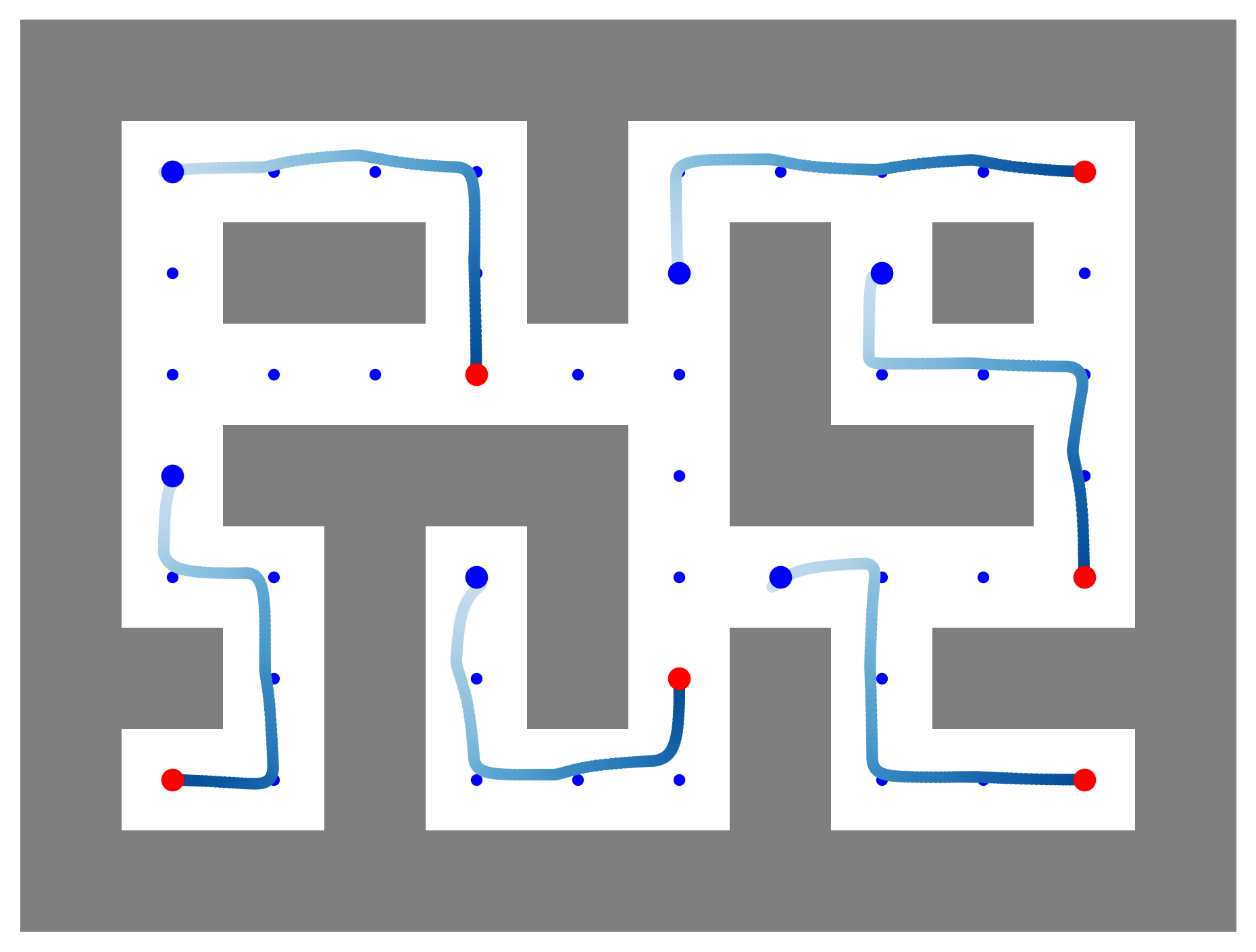}
        \caption*{(a) Large Maze}
    \end{minipage}
    \hfill
    \begin{minipage}{0.3\textwidth}
        \centering
        \includegraphics[width=\textwidth]{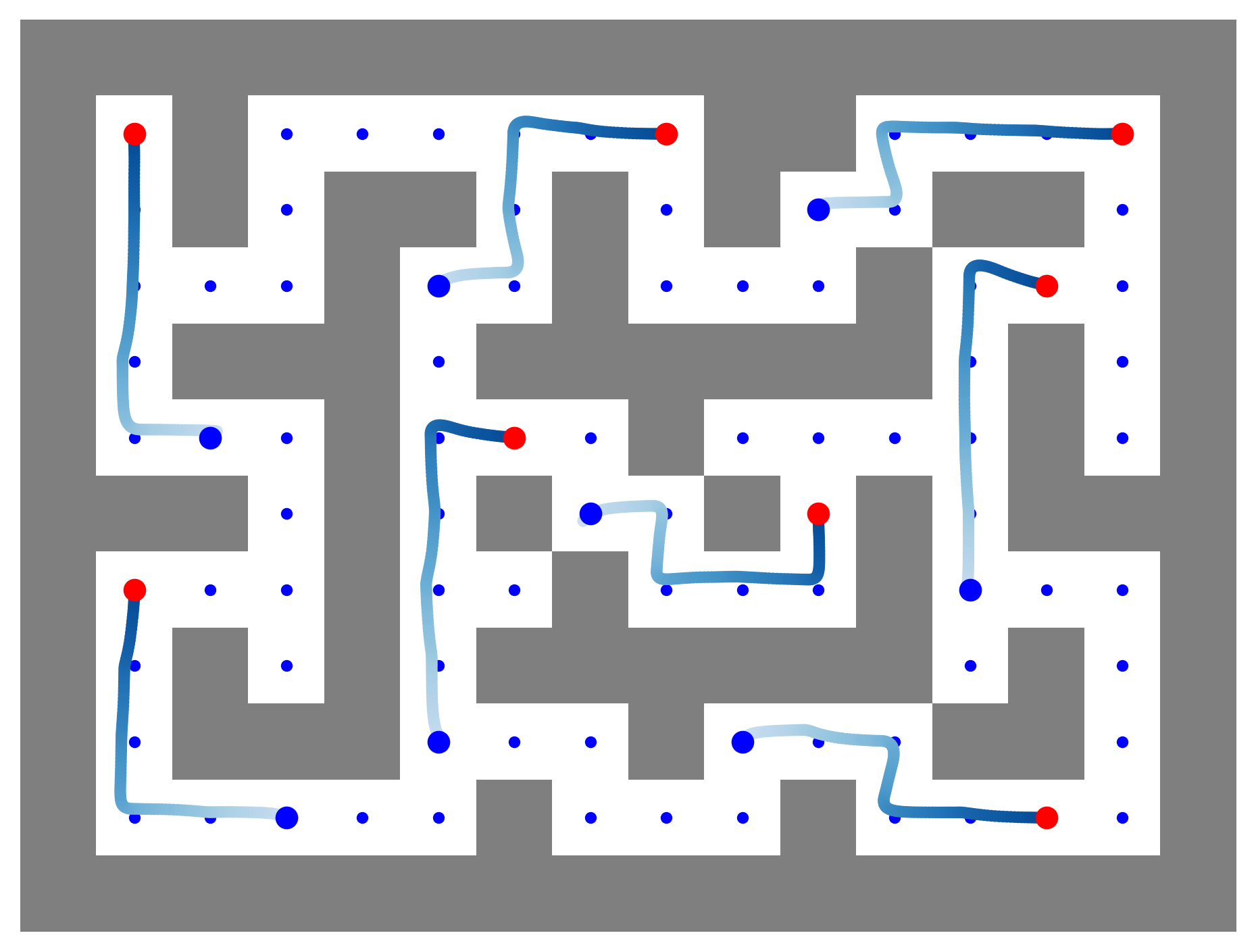}
        \caption*{(b) Giant Maze}
    \end{minipage}
    \hfill
    \begin{minipage}{0.3\textwidth}
        \centering
        \includegraphics[width=\textwidth]{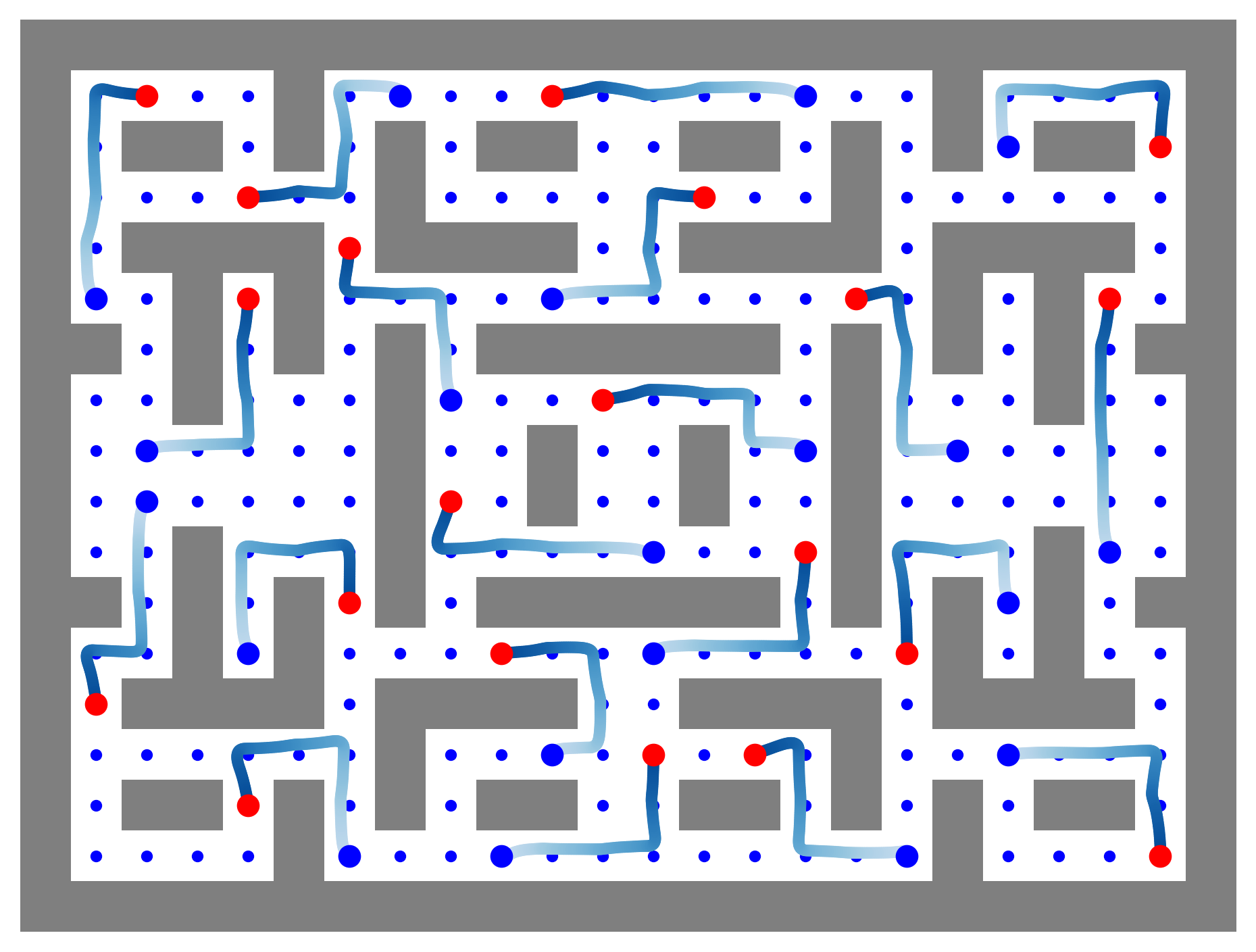}
        \caption*{(c) XXLarge Maze}
    \end{minipage}
    \caption{A few examples of base trajectories for each maze configuration. Each small blue point represents the center of a corresponding maze cell. We randomly sampled start-goal pairs with a maximum distance of five maze cells and followed D4RL to collect base trajectories, resulting in a total of one million transitions in the Base dataset.}
    \label{fig:base_trajectories}
\end{figure*}

\textbf{Maze Layouts.} Our experiments are conducted on 3 maze-layouts variying in lengths, we used the Large Maze ($9\times12$) from D4RL \cite{d4rl}, the Giant Maze ($12\times16$) from \cite{park2024ogbench}, and designed a new XXLarge Maze ($18\times24$) with a larger layout as shown in Figure \ref{fig:base_trajectories}. The sizes are measured in maze block cells.

\textbf{Data Collection.} For data collection, we randomly sample start and goal locations within five maze cells to ensure short trajectories. Following D4RL, we collected one million transitions for each Maze setting and we call it the base dataset, as depicted in Figure \ref{fig:base_trajectories}. Then, we run $R$ rounds of PTE starting from the short base dataset. 

\textbf{Training.} All models are trained with a fixed maximum planning horizon: $H=390$ for Large, $H=500$ for Giant, and $H=780$ for XXLarge. For HM-Diffuser, we use the hyperparameters as shown in Table \ref{tab:maze2d_hp}. For consistency, we set the maximum of the jump lengths for all environments to be $15$ to be equivalent to HD \cite{hd}. The jump counts are determined by the highest jump length $j_L$ and the planning horizon $H$.

\begin{table}[ht!]
\caption{\textbf{Maze2D HM-Diffuser Hyperparameters.}}
\small
\centering
\begin{adjustbox}{width=.6\linewidth}
\begin{tabular}{lclc}
\toprule 

 \multicolumn{1}{c}{\textbf{Environment}}   & Number of levels $L$ & jump lengths $j_\ell$ & jump counts $k_\ell$\\ \midrule

Large & $4$ & $(1, 8, 12, 15)$ & $26$\\
Giant & $5$& $(1,6,9,12,15)$ & $34$\\
XXLarge & $6$& $(1,6,8,10,12,15) $& $52$\\

\bottomrule
\end{tabular}
\end{adjustbox}
\label{tab:maze2d_hp}
\end{table}

\begin{figure}[ht!]
    \centering
    \includegraphics[width=0.8\textwidth]{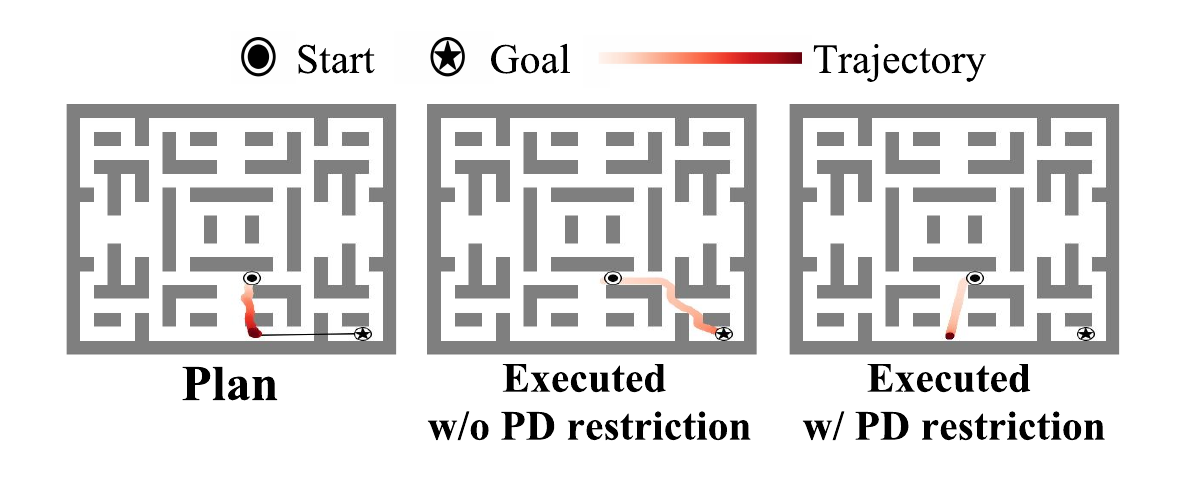}
    \vspace{-.5cm}
    \caption{\textbf{An illustrative example for the PD controller restriction.} This illustrates an example where the agent successfully reaches the goal even when the planning trajectory does not accurately lead to the goal.}
    \label{fig:app:pd_restriction}
\end{figure}

\textbf{Evaluation.} We report the mean and standard error over $N=200$ seeds, the performance is measured by the normalized scores, scaled by a factor of 100 \cite{d4rl, janner2022diffuser}. Each maze has a maximum number of steps per episode, $T=800$ for Large, $T=1000$ for Giant, and $T=1300$ for XXLarge. Following the Diffuser evaluation protocol, a proportional-derivative (PD) controller is used during evaluation to execute planned trajectories. However, we observed that the PD controller makes the goal reachable even when the planned trajectories are very poor (see to Figure \ref{fig:app:pd_restriction} for an illustrative example). To ensure that performance primarily reflects the model’s planning capabilities rather than controller interventions, we restrict the PD controller to be callable only when the agent are near the goal. Figure \ref{fig:app:pd_restriction} illustrates an example where the agent successfully reaches the goal even when the planning trajectory does not accurately lead to the goal.



\subsection{Linear PTE and Exponential PTE}
Table \ref{app:exp:maze2d_performance_exponential_datasetSize} shows the impact of the data size collected per round and PTE strategy on HM-Diffuser performance in Maze2D-XXLarge. Linear PTE shows steady improvements, reaching $89.9 \pm 8.2$ (Round 7, 4M). In contrast, Exponential PTE exhibits greater sensitivity to dataset size: smaller datasets lead to performance drops, whereas larger datasets (4M) achieve the highest score ($120.5 \pm 8.6$). These results highlight that Exponential PTE requires sufficient data to achieve optimal performance. We attribute this behavior to the fact that Exponential PTE exponentially increases trajectory lengths while keeping the dataset size fixed. As a result, the number of trajectories decreases, leading to reduced trajectory diversity for the source trajectory sampling, which in turn necessitates collecting more data per round.

\begin{table}[!t]
\caption{\textbf{Comparison of HM-Diffuser performance in the Maze2D-XXLarge environment, evaluating different PTE strategies.} The table presents results for varying rounds of PTE ($5$ and $7$) and differing amounts of data collected per round.}
\label{app:exp:maze2d_performance_exponential_datasetSize}
\small
\centering
\small
\centering
\begin{tabular}{crrrrrr}
\toprule 
\multicolumn{1}{c}{\multirow{3}{*}{\textbf{R}}} & \multicolumn{3}{c}{\textbf{Linear}} & \multicolumn{3}{c}{\textbf{Exponential}}    
\\ \cmidrule(l){2-4} \cmidrule(l){5-7}
\multicolumn{1}{c}{}        & \multicolumn{1}{c}{\textbf{1M}}     &\multicolumn{1}{c}{\textbf{2M}}    & \multicolumn{1}{c}{\textbf{4M}} & \multicolumn{1}{c}{\textbf{1M}}    & \multicolumn{1}{c}{\textbf{2M}} & \multicolumn{1}{c}{\textbf{4M}}  \\ \midrule

{5} & $68.9 \pm 7.2$ & $69.6 \pm 7.4$ & $70.1 \pm 7.7$& $53.4 \pm 6.7$ & $94.7 \pm 8.0$& $115.4 \pm 7.9$ \\ 
{7} & $82.1 \pm 8.5$ & $86.4 \pm 8.3$ & $89.9 \pm 8.2$ & $30.0 \pm 5.5$ & $51.3 \pm 6.7$ & $120.5 \pm 8.6$ \\ \bottomrule
\end{tabular}
\end{table}

\subsection{Additional results for Maze2D}

\begin{figure}[ht!]
    \centering
    \includegraphics[width=0.9\linewidth]{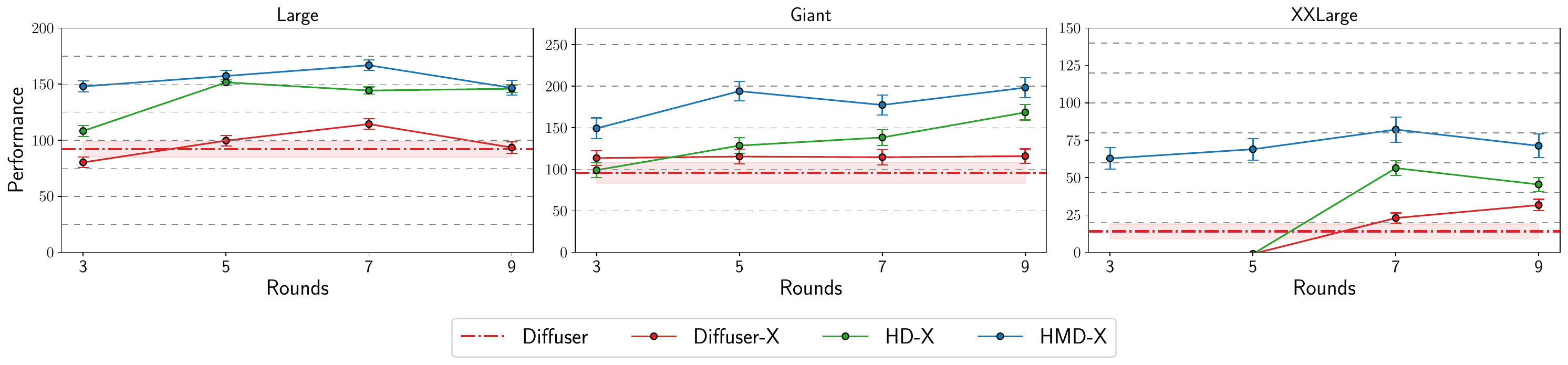}\\[0.5em]
    \includegraphics[width=0.9\linewidth]{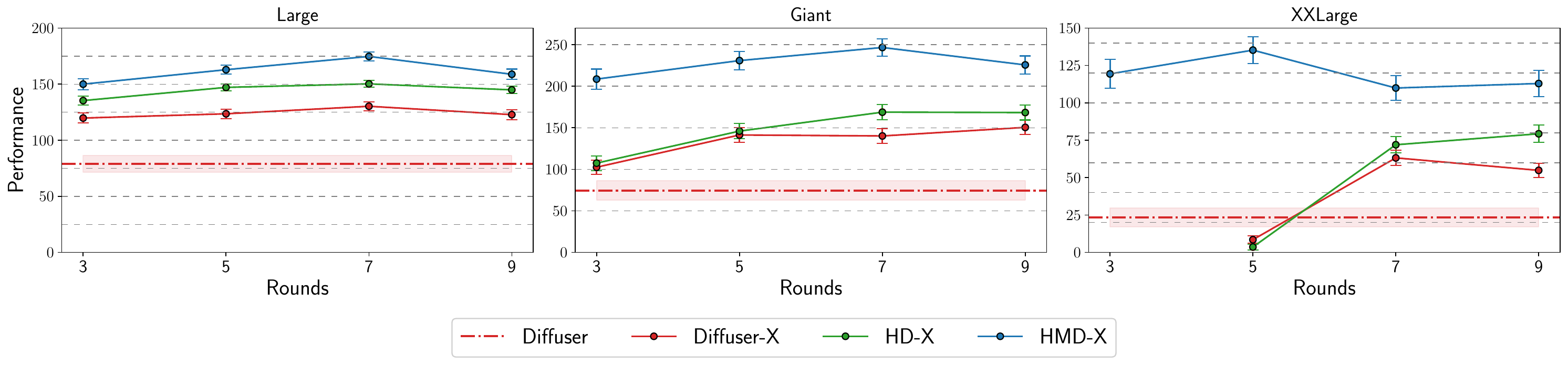}
    \caption{\textbf{Maze2D Performance.} Top: single-task (Maze2D). Bottom: multi-task (Multi2D). We compare the performance of baselines in various maze environments (Large, Giant, and XXLarge). The -X baselines are trained with the extended dataset consisting of the union of $R$ rounds of Linear PTE, while Diffuser is only trained with the short base dataset. We report the mean and the standard error over 200 planning seeds. HM-Diffuser-X consistently outperforms all baselines across all settings.}
    \label{fig:app:performance_rounds_maze2d_full}
\end{figure}

We present in Figure \ref{fig:app:performance_rounds_maze2d_full} the performance of baselines with the -X suffix, trained using various rounds of linear PTE ($3,5,7,9$). Training is conducted using the union of rounds; for example, performance shown in round 3, the models are trained with the dataset consists of trajectories collected from rounds 1, 2, and 3. Overall, performance improves with an increasing number of rounds. In the XXLarge setting, round 3 does not contain sufficiently long trajectories to train Diffuser-X or HD-X. However, HM-Diffuser-X, leveraging its multiscale planning horizons, is able to train using the available trajectories from round 3. A similar explanation applies to the poor performance of round 5 in XXLarge. The number of available trajectories with a planning horizon of $H = 780$ is limited, which affects the training of Diffuser-X and HD-X. Refer to Figure \ref{fig:app:PTE_histogram} for the trajectories lengths across PTE rounds.

\begin{figure}[ht!]
    \centering
    \includegraphics[width=0.6\textwidth]{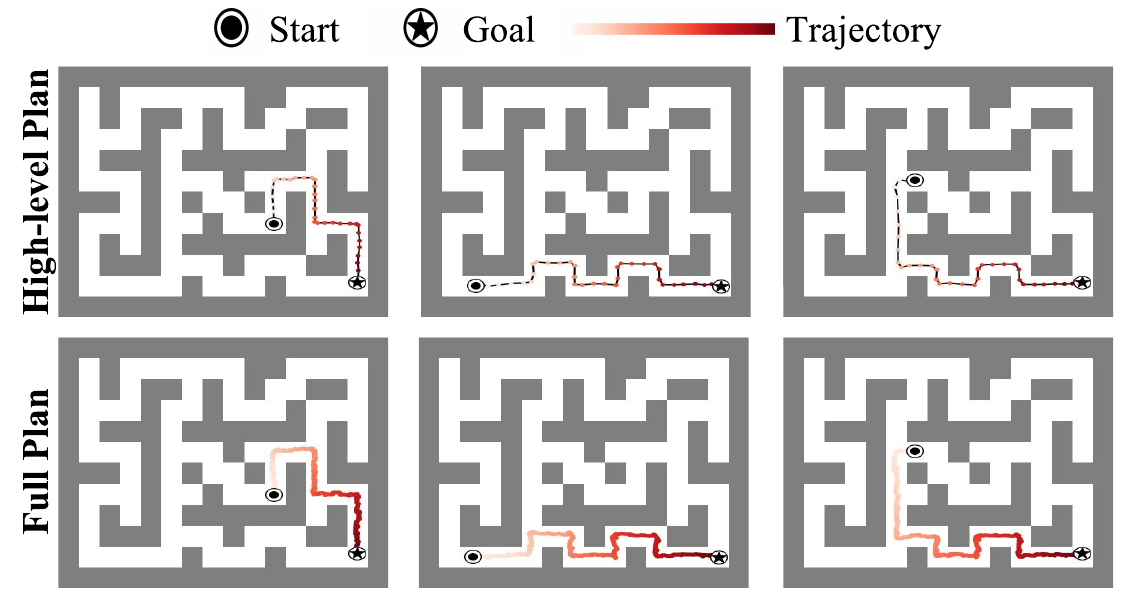}
    \caption{\textbf{Illustrative Examples of HM-Diffuser's Hierarchical Planning and Full Generated Plans.}  The high-level plan refers to the trajectory generated by HM-Diffuser when conditioned on the predicted level. The full plan represents the final output of HM-Diffuser after incorporating lower-level planning.}
    \label{fig:app:hier}
\end{figure}

Figure \ref{fig:app:hier} illustrates the high-level plans generated by HM-Diffuser when conditioned on the predicted level. Following this, HM-Diffuser is recursively called at lower levels to generate the full plan (see Section \ref{subsec:recursive_HMD}).

\begin{figure}[ht!]
    \centering
    \includegraphics[width=0.6\textwidth]{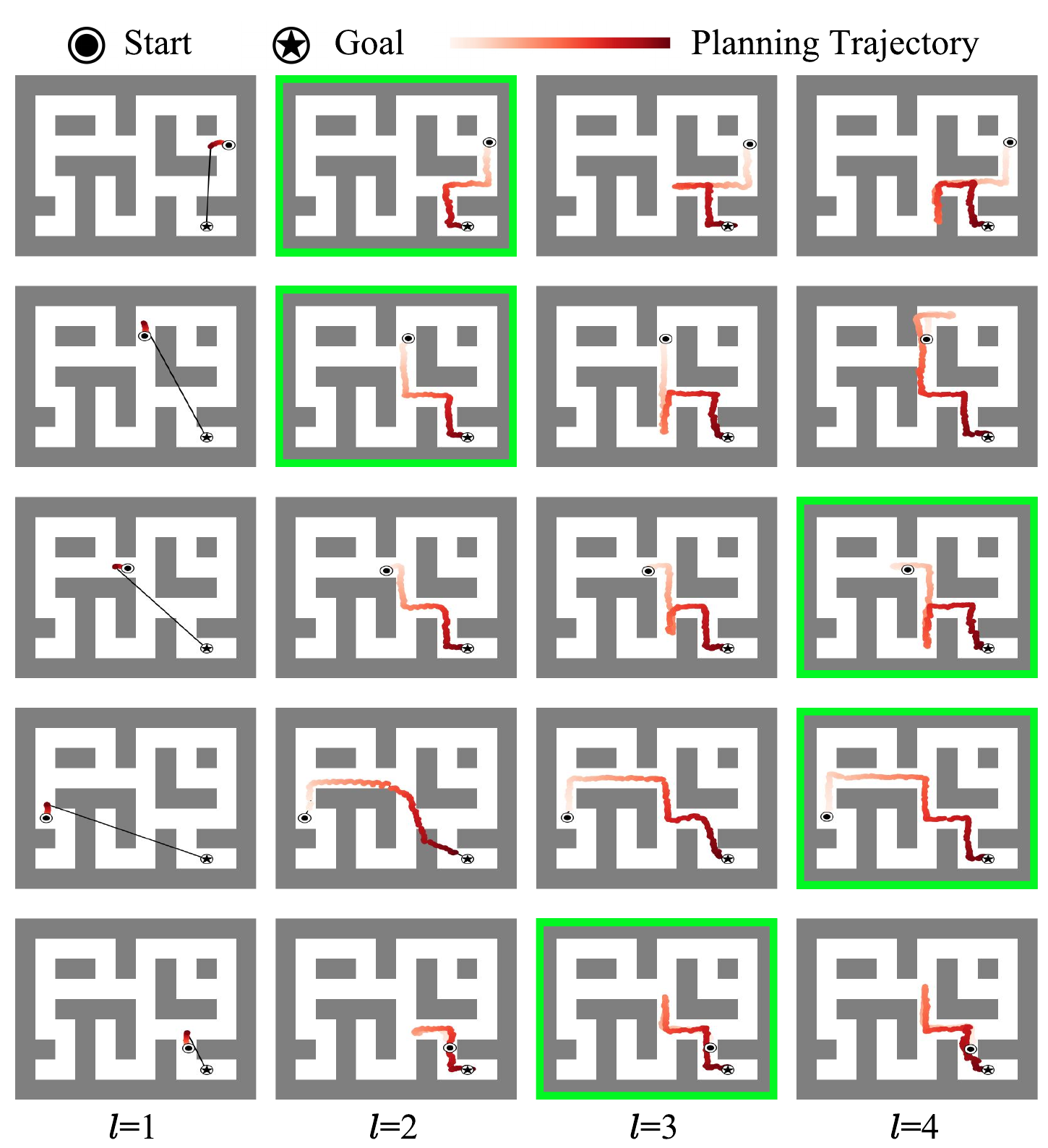}
    \caption{\textbf{Multiscale trajectory planning using HM-Diffuser with different levels ($l$)}, where the figure highlighted with a green border indicates the level predicted by the pondering depth predictor.}
    \label{fig:app:multiscale_planning}
\end{figure}

\begin{figure}[ht!]
    \centering
    \includegraphics[width=0.6\textwidth]{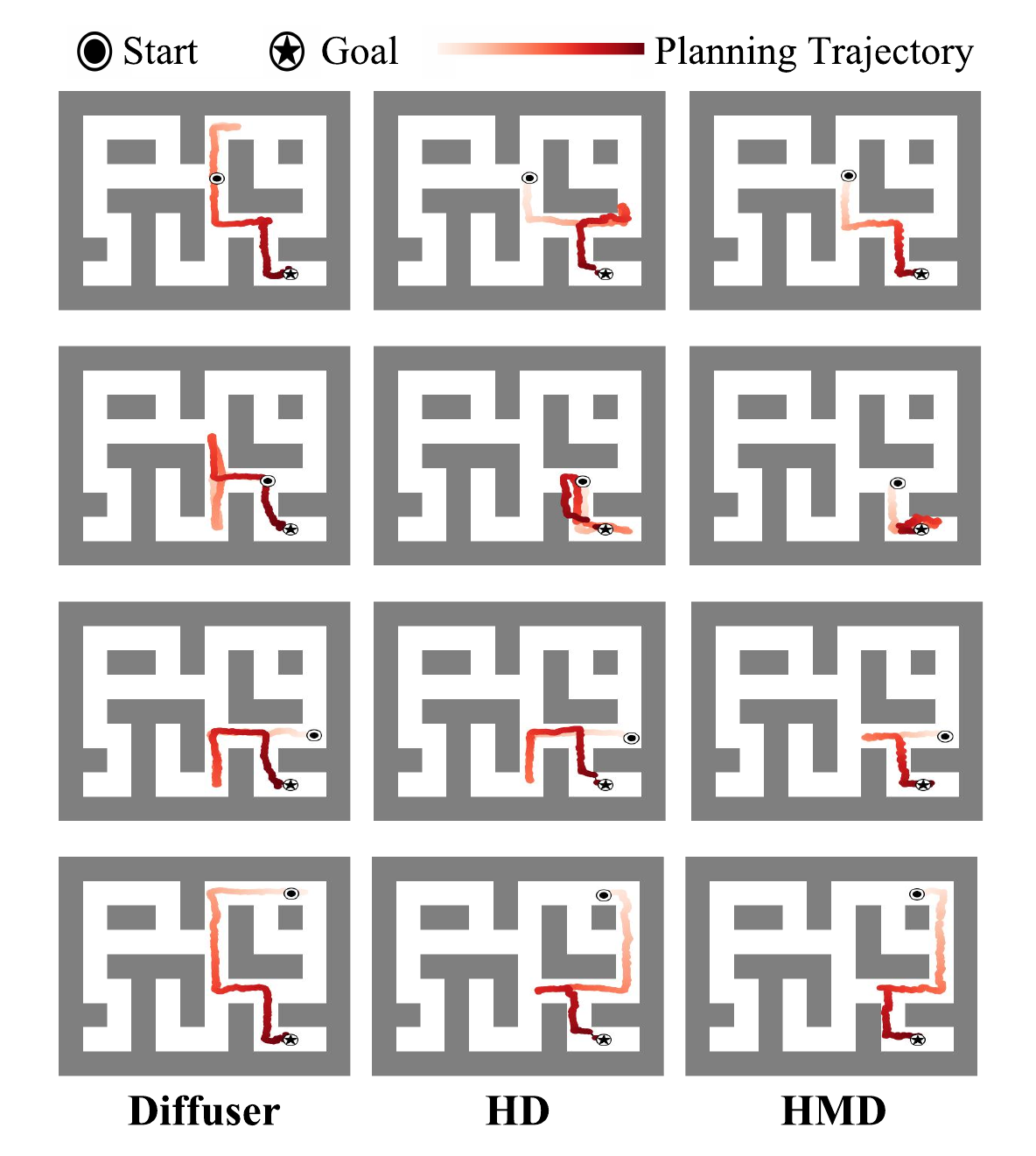}
    \caption{\textbf{Comparison of planning trajectories using different models (Diffuser, HD, and HM-Diffuser)} illustrating variations in trajectory planning efficiency and goal-reaching performance.}
    \label{fig:app:trajectory_planning_methods}
\end{figure}

\newpage



\newpage
\section{Progressive Trajectory Extension (PTE)} \label{app:PTE}

\begin{figure*}[ht!]
    \centering
    \includegraphics[width=0.9\textwidth]{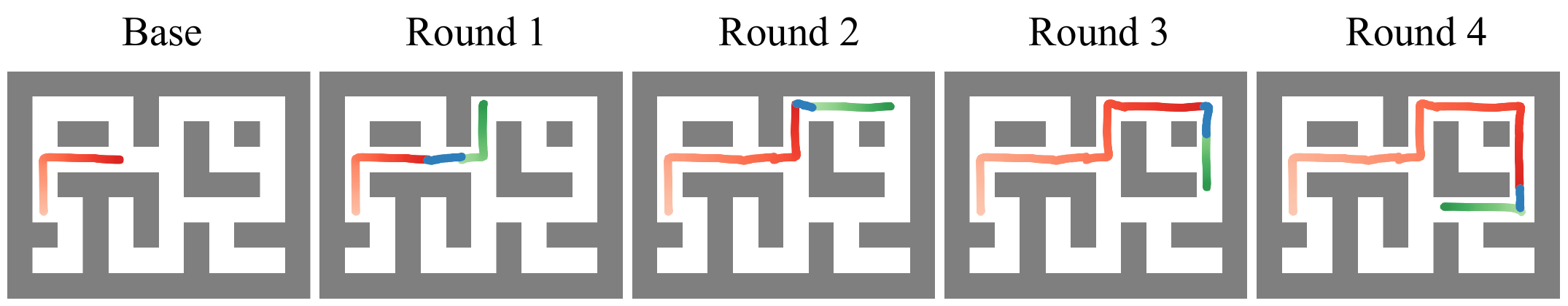}
    \caption{\textbf{Linear PTE over multiple rounds, starting from a single base trajectory.} The current trajectory is shown in red, the bridge in blue, and the target trajectory in green. As rounds progress, the current trajectory is stitched to a new stitchable target trajectory sampled from the base dataset}
    \label{fig:app:stitching_over_rounds}
\end{figure*}

\begin{figure}[ht!]
    \label{tab:stitching_length}
    \centering
    \includegraphics[width=0.8\textwidth]{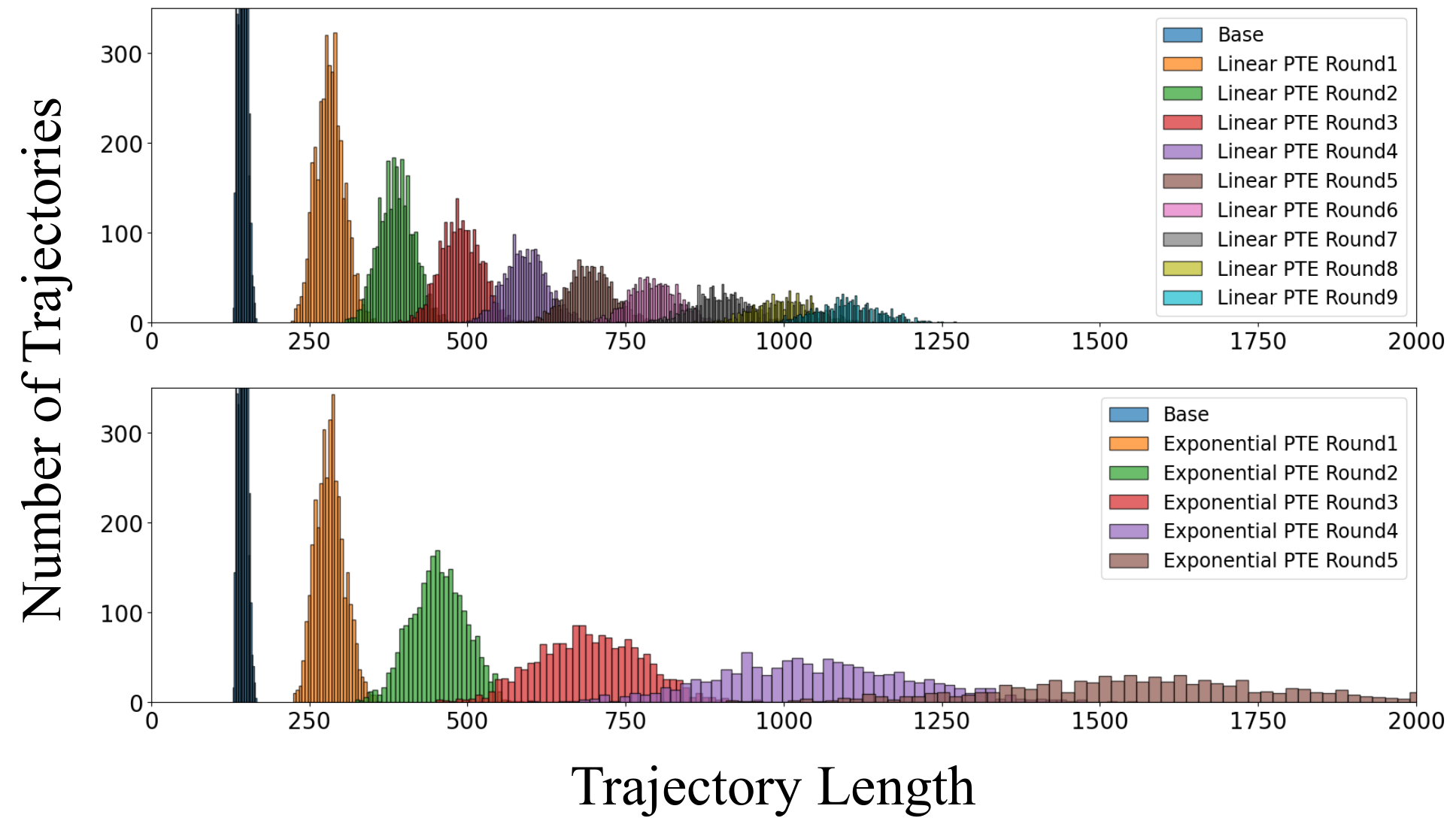}
    \caption{\textbf{Comparison of Trajectory Length Histograms Across PTE Rounds in Maze2D-XXLarge}. Exponential PTE shows a more rapid increase in trajectory length, with earlier rounds producing longer maximum trajectories compared to Linear PTE. Linear PTE, on the other hand, demonstrates a steadier, more gradual extension across rounds.}
    \label{fig:app:PTE_histogram}
\end{figure}

In this section, we first provide the pseudocode of our Progressive Trajectory Extension (PTE) process in algorithm \ref{alg:pte}. As discussed earlier, our PTE method allows flexible input datasets, thus enabling different stitching strategies. In algorithm \ref{alg:lienar_pte}, we highlighted the process of linear PTE, and the exponential PTE is depicted in algorithm \ref{alg:exponential_pte}. Additionally, Figure \ref{fig:app:stitching_over_rounds} shows an illustration of the linear PTE over multiple rounds, starting from a single base trajectory. Furthermore, the trajectory length histograms for both Linear PTE and Exponential PTE are shown in Figure \ref{fig:app:PTE_histogram}.

\subsection{Progressive Trajectory Extension (PTE) Pseudocode} 
\begin{algorithm}

\caption{Progressive Trajectory Extension}
\label{alg:pte} 
\begin{algorithmic}[1]
\State \textbf{Input:} Trained $p_\theta^{\text{stitcher}}$, Inverse Dynamic Model $f^a_\theta$, Reward Model $f^r_\theta$, Source Dataset $\cS$, Target Dataset $\cT$, Reachability Threshold $\delta$, Number of iterations $N$
\State \textbf{Output:} Stitched Dataset $D$
\\
\State Initialize $D \coloneqq \emptyset$
    
\For{$i = 1$ to $N$}
    \State\textbf{\# 1. Sampling a source trajectory, a bridge, and target candidates}
    \State Sample a source trajectory $\btau_s\sim \cS$ and a bridge trajectory $\btau_b \sim p_\ta^\text{stitcher}(\btau | s_0 = s_{\btau_s}^{\text{last}})$
    
    \State Sample a batch of target candidates $({\btau_t^{\scriptscriptstyle(1)}}, {\btau_t^{\scriptscriptstyle(2)}}, \dots, {\btau_t^{\scriptscriptstyle(c)}}) \subset \cT$

    \\
    \State\textbf{\# 2. Selecting a target trajectory}
    \State Filter out candidate trajectories and define the set of stitchable candidates $\mathcal{T}_{\text{stitch}}$ as:
    \[
        \mathcal{T}_{\text{stitch}} = \left\{ \btau_t^{\scriptscriptstyle(j)} \,\middle|\, 
            \min_{m, n} \text{dist}(s_t^{j, m}, s_b^{}) \leq \delta
        \right\}
    \]
    \State where $s_t^{j, m}$ denotes the $m$-th state of $\btau_t^{\scriptscriptstyle(j)}$, and $s_b^{n}$ denotes the $n$-th state of $\btau_b$.

    \State Randomly sample a target trajectory $\btau_t \sim \cT_{\text{stitch}}$
    \\
    \State\textbf{\# 3. Stitching the source and the target}
    \State Set $k \coloneqq \arg \min_{j} \text{dist}(s_t^j, s_b^n)$
    \State Re-sample the bridge trajectory \smash{$\btau_b \sim p_\ta^\text{stitcher}(\btau | s_0 = s_{\btau_s}^{\text{last}}, \  s_{\text{-k:}} = (s_t^0, s_t^1, \dots, s_t^k))$}
    \State where $s_t^{j}$ denotes the $j$-th state of $\btau_t$.

    \State Initialize $\btau_{b}^{\prime} \coloneqq \emptyset$
    \For{$j = 0$ to $k-1$}
        \State Predict action using the inverse dynamics model: $a_{j} \coloneqq f^a_\theta(s_{\btau_b}^{\scriptscriptstyle(j)}, s_{\btau_b}^{\scriptscriptstyle(j+1)})$
        \State Predict reward using the reward model: $r_{j} \coloneqq f^r_\theta(s_{\btau_b}^{\scriptscriptstyle(j)}, a_{j})$
        \State Append $\{s_{\btau_b}^{\scriptscriptstyle(j)}, a_{j}, r_{j}\}$ to $\btau_{b}^{\prime}$
    \EndFor
    
    \State Get $\btau^\text{new} = \btau_s \mathbin\Vert \btau_b^{\prime} \mathbin\Vert \btau_t$
    \State Update $D \coloneqq D \cup \btau^\text{new}$
\EndFor
\\
\State \textbf{Return:} Extended Dataset $D$
\end{algorithmic}
\end{algorithm}

\vspace{-.3cm}

\begin{algorithm}
\caption{Linear PTE}
\label{alg:lienar_pte}
\begin{algorithmic}[1]
    \State \textbf{Input:} Trained $p_\theta^{\text{stitcher}}$, Inverse Dynamic Model $f^a_\theta$, Reward Model $f^r_\theta$, Reachability Threshold $\delta$, Previous Round Dataset $\cD_{r-1}$, Base Dataset $\cD_0$, Number of iterations $N$
    \State \textbf{Output:} Stitched Dataset $D_r$

    Use Algorithm\ref{alg:pte} with \red{$\cS = \cD_{r-1}$,  $\cT = \cD_0$}
    
    \State \textbf{Return:} Extended Dataset $D_r$
\end{algorithmic}
\end{algorithm}
\vspace{-.3cm}

\begin{algorithm}
\caption{Exponential PTE}
\label{alg:exponential_pte}
\begin{algorithmic}[1]
    \State \textbf{Input:} Trained $p_\theta^{\text{stitcher}}$, Inverse Dynamic Model $f^a_\theta$, Reward Model $f^r_\theta$, Reachability Threshold $\delta$, Previous Round Dataset $\cD_{r-1}$, Number of iterations $N$
    \State \textbf{Output:} Stitched Dataset $D_r$

    Use Algorithm\ref{alg:pte} with \red{$\cS = \cT = \cD_{r-1}$}
    
    \State \textbf{Return:} Extended Dataset $D_r$
\end{algorithmic}
\end{algorithm}
\vspace{-.3cm}

\newpage
\subsection{Planning with Recursive HM-Diffuser}
We present the planning pseudocode with our proposed recursive HM-Diffuser in algorithm \ref{app:alg_replan}
.
\begin{algorithm}
\caption{Planning with Recursive HM-Diffuser - Replanning}
\label{app:alg_replan}
\begin{algorithmic}[1]
\State \textbf{Input:} HM-Diffuser $p_\theta$, Evaluation Environment \textit{env}, Inverse Dynamic $f^a_\theta$, Number of Levels $L$, Jump Count $K=\{k_\ell\}^L$

\Statex
\State $s_0$ = \textit{env}.init()
\Statex \Comment{Reset the environment.}
\State done = False
\While{not done}
    \For{$\ell$ in $L, \dots, 1$}
        \If{$\ell == L$}
            \State $\tau_\ell^g = \{g_\ell^{\scriptscriptstyle(0)}, \dots, g^{\scriptscriptstyle(k_\ell)}_\ell\} \gets p_\theta(\btau | \ell, g_\ell^{\scriptscriptstyle(0)} = s_0)$
            \Statex \Comment{Sample a subgoal plan given start.}
        \Else
            \State $\tau_\ell^g =  \{g_\ell^{\scriptscriptstyle(0)}, \dots, g_{k_\ell}^{\scriptscriptstyle(l)}\} \gets p_\theta(\btau | \ell, g_\ell^{\scriptscriptstyle(0)} = s_0, g_\ell^{\scriptscriptstyle(k_\ell)} = g_{\ell+1}^{\scriptscriptstyle(1)})$
            \Statex \Comment{Refine plans given subgoals from one layer above.}
        \EndIf
    \EndFor
    \State Extract the first two states, $s_0, s_{1}=g_1^{\scriptscriptstyle(0)}$, from the first layer plan $\tau_1^g$
    \State Obtain action $a = f^a_\theta(s_0, s_{1})$
    \State Execute action in the envirionment $s, \text{done} = env.step(a) $
\EndWhile

\end{algorithmic}
\end{algorithm}

\begin{algorithm}
\caption{Recursive HM-Diffuser Training}
\label{alg:recursive_hmd_training}
\begin{algorithmic}[1]
\State \textbf{Input:} Recursive HM-Diffuser $p_\theta$, Inverse Dynamic $f^a_\theta$, number of levels $L$, Reward Model $f^r_\theta$, Jumpy Step Schedule $J = \{j_0, \dots, j_L\}$, Training Dataset $\mathcal{D}$

\While{not done}
    \State Sample a batch of trajectory from dataset $\btau = \{s_t, a_t, r_t\}^{t+h} \sim \mathcal{D}$
    \State Sample a level $\ell \sim \text{Unifrom[0, \dots, L]}$
    \State Obtain the sparse trajectory for level $\ell$: $\btau^\ell = (g_\ell^{\scriptscriptstyle(0)}, \dots, g_\ell^{\scriptscriptstyle(k_\ell)})$
    \State Train HM-Diffuser with Equation 4
    \State Train inverse dynamics $f^a_\theta$
    \State Train reward model $f^r_\theta$
\EndWhile

\end{algorithmic}
\end{algorithm}


        





\subsection{Limits of PTE}\label{app:limit_pte}
It would be interesting to test the limits of our proposed PTE process. Intuitively, introducing a generative model during the extension introduces noise, making it straightforward to hypothesize that with successive rounds of extension, the planner might struggle to generate plausible plans due to the noisy dataset. We report our results for the Kitchen in Table \ref{app:performance_kitchen_detail} and for Gym-MuJoCo in Table \ref{app:performance_dmc_detail}. To better understand the quality of our extended dataset, we include results from Diffuser trained on short segments of equivalent length from the original dataset (denoted as ‘No PTE’) for comparison. As shown in the tables, the performance of all models declines with more rounds of PTE, except for one exception in the Walker2d-MedExpert task, where HD-X and HM-Diffuser-X achieved better performance after two rounds of PTE. Diffuser’s performance drops sharply as the planning horizon extends, whereas HD and HM-Diffuser, benefiting from their hierarchical structure, remain more stable in comparison. We acknowledge this limitation and leave further investigation for future work.
\begin{table*}[ht!]
\small
\centering
\caption{Kitchen results from more rounds of PTE.}
\label{app:performance_kitchen_detail}
\begin{adjustbox}{width=\linewidth}
\begin{tabular}{cccccccccccccc}
\toprule
\multirow{3}{*}{Task}
& length=10 & \multicolumn{4}{c}{length = 20}       & \multicolumn{4}{c}{length = 40}           & \multicolumn{4}{c}{length = 80}    \\ \cmidrule(l){2-2} \cmidrule(l){3-6} \cmidrule(l){7-10} \cmidrule(l){11-14}
& Base Trajectory & \multicolumn{3}{c}{Round 1 PTE}       & No PTE    & \multicolumn{3}{c}{Round 2 PTE}      & No PTE      & \multicolumn{3}{c}{Round 3 PTE}    & No PTE  \\ \cmidrule(l){2-2} \cmidrule(l){3-5} \cmidrule(l){6-6} \cmidrule(l){7-9} \cmidrule(l){10-10} \cmidrule(l){11-13} \cmidrule(l){14-14}
& Diffuser & Diffuser-X  & HD-X        & HM-Diffuser-X      & Diffuser      & Diffuser-X & HD-X    & HM-Diffuser-X       & Diffuser   & Diffuser-X  & HD-X      & HM-Diffuser-X       & Diffuser     \\ \midrule
Kitchen-Partial        &$41.7 \pm 3.2$ & $43.3 \pm 5.5$ & $\mathbf{56.7} \pm 5.8$ & $56.7 \pm 5.3$ &  $35.8 \pm 2.6$ & $23.3 \pm 6.3$ & $\mathbf{41.7} \pm 2.9$ & $\mathbf{50.0} \pm 3.9$ & $26.7 \pm 2.9$ & $13.3 \pm 3.1$ & $40.0 \pm 3.1$ & $\mathbf{45.0} \pm 4.1$ & $25.8 \pm 2.8$ \\
Kitchen-Mixed          &$45.8 \pm 3.1$ & $48.3 \pm 4.7$ & $53.3 \pm 3.1$         & $\mathbf{61.7} \pm 3.1$   &  $40.8 \pm 2.8$ & $40.0 \pm 5.4$    & $\mathbf{50.0} \pm 3.9$       & $50.0 \pm 4.6$  & $30.8 \pm 4.2$ & $18.3 \pm 4.2$  & $46.7 \pm 4.5$        &$\mathbf{53.3} \pm 5.0$         & $32.5 \pm 3.9$ \\ \midrule
\textbf{Kitchen Average} & $43.8$ &$45.8$ & $55.0$ &$59.2$ &$38.3$ &$31.7$ &$48.9$ &$50.0$ &$28.8$ &$15.8$ &$43.4$ &$49.2$ &$29.2$ \\
\bottomrule

\end{tabular}
\end{adjustbox}
\end{table*}

\begin{table*}[ht!]
\small
\centering
\caption{Gym-MuJoCo results from more rounds of PTE.}
\label{app:performance_dmc_detail}
\begin{adjustbox}{width=\linewidth}
\begin{tabular}{cccccccccc}
\toprule
\multirow{3}{*}{Task}
& length=10 & \multicolumn{4}{c}{length = 20}       & \multicolumn{4}{c}{length = 40}             \\ \cmidrule(l){2-2} \cmidrule(l){3-6} \cmidrule(l){7-10} 
& Base Trajectory & \multicolumn{3}{c}{Round 1 PTE}       & No PTE    & \multicolumn{3}{c}{Round 2 PTE}      & No PTE     \\ \cmidrule(l){2-2} \cmidrule(l){3-5} \cmidrule(l){6-6} \cmidrule(l){7-9} \cmidrule(l){10-10} 
                    & Diffuser      & Diffuser-X     & HD-X                     & HM-Diffuser-X         & Diffuser          & Diffuser-X     & HD-X          & HM-Diffuser-X          & Diffuser    \\ \midrule
Walker2d-MedReplay  &$22.8 \pm 2.7$ & $20.1 \pm 4.3$ & $\mathbf{30.2} \pm 5.9$  &$29.6 \pm 4.8$ & $24.9 \pm 4.4$    & $19.5 \pm 2.6$ &$\mathbf{27.5} \pm 2.6$ & $22.7 \pm 2.4$ & $22.4 \pm 4.2$  \\
Walker2d-Medium     &$58.1 \pm 5.6$ & $62.6 \pm 6.4$ & $66.5 \pm 4.3$           &$\mathbf{72.7} \pm 2.5$ & $24.9 \pm 6.8$    & $41.0 \pm 7.8$ &$\mathbf{55.3} \pm 6.1$ & $48.5 \pm 5.4$ & $23.7 \pm 6.3$  \\ 
Walker2d-MedExpert  &$82.3 \pm 4.6$ & $80.3 \pm 3.7$ & $\mathbf{80.8} \pm 2.9$  &$79.3 \pm 2.3$ & $77.7 \pm 10.4$   & $65.1 \pm 8.3$ &$\mathbf{85.1} \pm 3.9$ & $84.7 \pm 4.6$ & $61.7 \pm 6.8$  \\ \midrule
\textbf{Walker2d Average}  & $54.4$ & $54.3$ & $59.2$ & $\mathbf{60.5}$ & $42.5$ &$41.9$ &$56.0$ &$52.0$ &$35.9$ \\ \midrule
Hopper-MedReplay    &$18.7 \pm 3.0$ & $34.5 \pm 6.2$ & $22.5 \pm 3.1$  &$\mathbf{37.3} \pm 4.2$ & $29.1 \pm 4.4$  & $22.9 \pm 4.1$ &$\mathbf{36.5} \pm 4.9$ & $31.6 \pm 3.3$ & $25.7 \pm 3.8$  \\
Hopper-Medium       &$45.6 \pm 1.9$ & $44.3 \pm 3.5$ & $44.1 \pm 2.8$  &$\mathbf{44.9} \pm 3.5$ & $50.5 \pm 3.8$  & $\mathbf{44.5} \pm 2.2$ &$34.9 \pm 2.8$ & $42.1 \pm 2.6$ & $42.7 \pm 2.1$  \\ 
Hopper-MedExpert    &$61.4 \pm 8.4$ & $\mathbf{74.9} \pm 8.0$ & $67.9 \pm 7.7$  &$74.3 \pm 9.0$ & $61.3 \pm 7.4$  & $48.7 \pm 4.8$ &$\mathbf{46.4} \pm 5.3$ & $\mathbf{60.8} \pm 8.5$ & $52.2 \pm 3.5$  \\ \midrule
\textbf{Hopper Average}  & $41.9$   & $51.2$         & $44.8$ & $\mathbf{52.2}$        & $46.6$          & $38.7$         &$39.3$ &$\mathbf{44.8}$ &$40.2$ \\

\bottomrule

\end{tabular}
\end{adjustbox}
\end{table*}

.

\section{Extendable Franka Kitchen and Extendable Gym-MuJoCo}\label{app:additional_kitchen_dmc}

\textbf{Extended Dataset.} To generate our extended dataset, we first construct the base dataset, which consists of short segments extracted from the standard D4RL dataset. Specifically, we partition the original D4RL offline dataset into non-overlapping segments of length 10. Next, we train a stitcher, implemented as a standard Diffuser, on this base dataset. To extend the trajectories, we uniformly sample a source trajectory and bridge it to a target trajectory, as described in Section \ref{sec:PTE}. This process extends the trajectory lengths, which enables planning beyond the horizon of training dataset for diffusion-based planners. We generate a dataset of the same size as the standard offline dataset for Gym-MuJoCo and three times the size for Kitchen.

\textbf{Training.} We follow the training protocal as in Diffuser and HD. We deploy a two-layer HM-Diffuser on Kitchen and Gym-MuJoCo, with $j_2 = 4$ and $j_1 = 1$. The planning horizon the set to be the minimum length of the extended trajectory, i.e. $H=20$ for round 1 extension, $H=40$ for round 2 extension, and $H=80$ for round 3 extension.

\section{Additional Results}
\textbf{Effects of Planning Horizon on Diffuser.} We provide the results from Diffuser on dataset with varied segment length in Table \ref{app:performance_dmc_horizon}. The Infinite denotes the original setting, where the trajectory return is computed till the end of the episode. 
\begin{table}[t!]
\small
\centering
\caption{Gym-MuJoCo performance with varied planning horizon.}
\label{app:performance_dmc_horizon}
\begin{tabular}{llrrr}

\toprule 
\multicolumn{2}{c}{\textbf{Environment}} & \textbf{H5}            & \textbf{H10}       & \textbf{Infinite}    \\ \midrule      
Walker2D            & MedReplay          & $21.2 \pm 8.1$         &$22.8 \pm 2.7$      & $76.1 \pm 5.0$       \\
Walker2D            & Medium             & $60.2 \pm 3.2$         &$58.1 \pm 5.6$      & $81.8 \pm 0.5$       \\
Walker2D            & MedExpert          & $75.9 \pm 4.3$         &$82.3 \pm 4.6$      & $106.5 \pm 0.2$      \\ \midrule
\multicolumn{2}{c}{\textbf{Walker2D Average}}   &$52.4$           &$54.4$              & $88.1$              \\ \midrule
Hopper              & MedReplay          & $24.3 \pm 5.0$         &$18.7 \pm 3.0$      & $93.6 \pm 0.4$       \\
Hopper              & Medium             & $43.9 \pm 1.8$         &$45.6 \pm 1.9$      & $74.3 \pm 1.4$       \\
Hopper              & MedExpert          & $59.2 \pm 6.6$         &$61.4 \pm 8.4$      & $103.3 \pm 1.3$      \\ \midrule
\multicolumn{2}{c}{\textbf{Halfcheetah Average}}   &$42.4$        &$41.9$              & $90.4$                  \\ \midrule
Halfcheetah         & MedReplay          & $35.3 \pm 3.7$         &$34.8 \pm 4.8$      & $37.7 \pm 0.5$       \\
Halfcheetah         & Medium             & $39.1 \pm 3.2$         &$45.2 \pm 3.3$      & $42.8 \pm 0.3$       \\
Halfcheetah         & MedExpert          & $72.4 \pm 6.8$         &$69.4 \pm 7.5$      & $42.8 \pm 0.3$      \\ \midrule
\multicolumn{2}{c}{\textbf{Halfcheetah Average}}   &$48.9$        &$49.8$              & $56.5$              \\ 

\bottomrule

\end{tabular}
\end{table}

\textbf{Performance on Standard Benchmark.} We provide the results on standard D4RL benchmark in Table \ref{app:performance_standard_benchmark}.
\begin{table}[t!]
\small
\centering
\caption{\textbf{HM-Diffuser on Standard Benchmark.} HM-Diffuser noticeably outperforms HD and Diffuser on the Maze2D tasks and performs comparably to HD on Kitchen and Gym-MuJoCo tasks.}
\label{app:performance_standard_benchmark}
\begin{tabular}{llrrr}

\toprule 
\multicolumn{2}{c}{\textbf{Environment}} & \textbf{Diffuser}      & \textbf{HD}         & \textbf{HM-Diffuser}    \\ \midrule      
Maze2D                  & Large                 & $128.6 \pm 2.9$   & $155.8\pm 2.5$           & $\mathbf{177.3} \pm 3.89$\\
Maze2D                  & Giant                 & $86.9 \pm 8.4 $   & $173.9 \pm 8.7$ & $\mathbf{209.4} \pm 11.9$\\
Maze2D                  & XXLarge               & $61.9 \pm 4.6$  & $137.1\pm 4.4$           & $\mathbf{146.7} \pm 9.1$\\ \midrule
\multicolumn{2}{c}{\textbf{Sing-task Average}}    & $92.5$ & $155.6$ & $\mathbf{177.8}$                    \\ \midrule
Multi2D             & Large              & $132.1 \pm 5.8$ & $165.5 \pm 0.6$ & $\mathbf{181.3} \pm 4.1$      \\ 
Multi2D             & Giant              & $131.7 \pm 8.9$ & $181.3 \pm 8.9$ & $\mathbf{258.9} \pm 10.4$     \\ 
Multi2D             & XXLarge              & $86.7 \pm 5.6$ & $150.3 \pm 4.2$ & $\mathbf{209.5} \pm 8.8$      \\ \midrule
\multicolumn{2}{c}{\textbf{Multi-task Average}}  & $116.8$ & $165.7$ & $\mathbf{216.6}$              \\ \midrule
Walker2D            & MedReplay          & $70.6 \pm 1.6$         &$\mathbf{84.1} \pm 2.2$      & $80.7 \pm 3.2$       \\
Walker2D            & Medium             & $79.9 \pm 1.8$         &$\mathbf{84.0} \pm 0.6$      & $82.2 \pm 0.6$       \\
Walker2D            & MedExpert          & $106.9 \pm 0.2$        &$107.1 \pm 0.1$              & $\mathbf{107.6 \pm 0.7}$      \\ \midrule
\multicolumn{2}{c}{\textbf{Walker2D Average}}   &$85.8$        &$\mathbf{91.7}$              & $89.9$              \\ \midrule
Halfcheetah         & MedReplay          & $37.7 \pm 0.5$         &$38.1 \pm 0.7$               & $\mathbf{41.1} \pm 0.2$       \\
Halfcheetah         & Medium             & $42.8 \pm 0.3$         &$\mathbf{46.7} \pm 0.2$      & $44.9 \pm 1.1$       \\
Halfcheetah         & MedExpert          & $88.9 \pm 0.3$         &$92.5 \pm 1.4$               & $90.6 \pm 1.3$      \\ \midrule
\multicolumn{2}{c}{\textbf{Halfcheetah Average}}   &$56.5$        &$\mathbf{59.1}$                 & $58.9$              \\ \midrule
Hopper              & MedReplay          & $93.6 \pm 0.4$         &$94.7 \pm 0.7$               & $\mathbf{95.5} \pm 0.9$       \\
Hopper              & Medium             & $74.3 \pm 1.4$         &$\mathbf{99.3} \pm 0.3$      & $99.2 \pm 0.1$       \\
Hopper              & MedExpert          & $103.3 \pm 1.3$         &$115.3 \pm 1.1$               & $113.6 \pm 2.7$      \\ \midrule
\multicolumn{2}{c}{\textbf{Hopper Average}}   &$90.4$        &$\mathbf{103.1}$                 & $102.8$              \\ \midrule
FrankaKitchen       & Partial            & $55.0 \pm 10.0$        &$\mathbf{73.3} \pm 1.4$      & $71.5 \pm 2.0$      \\
FrankaKitchen       & Mixed              & $58.3 \pm 4.5$         &$71.5 \pm 2.3$               & $\mathbf{71.8} \pm 3.3$   \\ \midrule
\multicolumn{2}{c}{\textbf{FrankaKitchen Average}}   &$56.7$        &$\mathbf{72.4}$            & $71.7$              \\ 
\bottomrule

\end{tabular}
\end{table}

\textbf{Addtional Hierarchical Baseline Results on Extendable Planning Problem.} DiffuserLite \citep{dong2024diffuserlite}, a variant hierarchical planner, lacks data augmentation or trajectory stitching capability, making it fundamentally limited in settings that require expanded planning. As shown in Table \ref{app:exp:hmd_diifuserlite}, DiffuserLite alone performs poorly compared with HM-Diffuser-X. As well be shown in Table \ref{app:exp:pte_diffsttich}, naively combining DiffuserLite and DiffStitch is insufficient. This combination performs worse than our proposed framework because it inherits the coverage limitations of DiffStitch.

\begin{table}[t!]
\small
\centering
\caption{Our proposed framework works the best in the extendable planning problem.}
\label{app:exp:hmd_diifuserlite}
\begin{tabular}{llrrr}
\toprule 
\multicolumn{2}{c}{\textbf{Environment}}        & \textbf{DiffStitch}    &\textbf{DiffuserLite-Unet}    & \textbf{HM-Diffuser-X}    \\ \midrule
Kitchen & Partial-v0 & $21.7 +- 2.1$ & 11.7 +- 3.1 & \textbf{56.7} +- 5.3 \\
Kitchen & Mixed-v0 & 18.3 +- 2.7 & 8.3 +- 2.9 & \textbf{61.7} +- 3.1 \\ \midrule
\multicolumn{2}{c}{\textbf{Kitchen Average}} & 20.0 & 10.0 & \textbf{59.2} \\ \midrule
Walker2d & MedReplay & \textbf{34.1} +- 6.1 & 7.2 +- 0.9 & 29.6 +- 4.8 \\
Walker2d & Medium & 31.3 +- 7.2 & 31.4 +- 3.7 & \textbf{72.7} +- 2.5 \\
Walker2d & MedExpert & 0.0 +- 0.0 & 54.3 +- 4.2 & \textbf{79.3} +- 2.3 \\ \midrule
\multicolumn{2}{c}{\textbf{Walker2d Average}} & 21.8 & 30.9 & \textbf{60.5} \\ \midrule
Hopper & MedReplay & 24.6 +- 3.8 & 14.4 +- 1.0 & \textbf{37.3} +- 4.8 \\
Hopper & Medium & 1.0 +- 0.0 & 43.9 +- 1.8 & \textbf{44.9} +- 3.5 \\
Hopper & MedExpert & 0.0 +- 0.0 & 68.1 +- 6.0 & \textbf{74.3} +- 9.0 \\ \midrule
\multicolumn{2}{c}{\textbf{Hopper Average}} & 8.5 & 42.1 & \textbf{52.1} \\
\bottomrule
\end{tabular}
\end{table}

\begin{figure}[t!]
    \centering
    \includegraphics[width=0.8\textwidth]{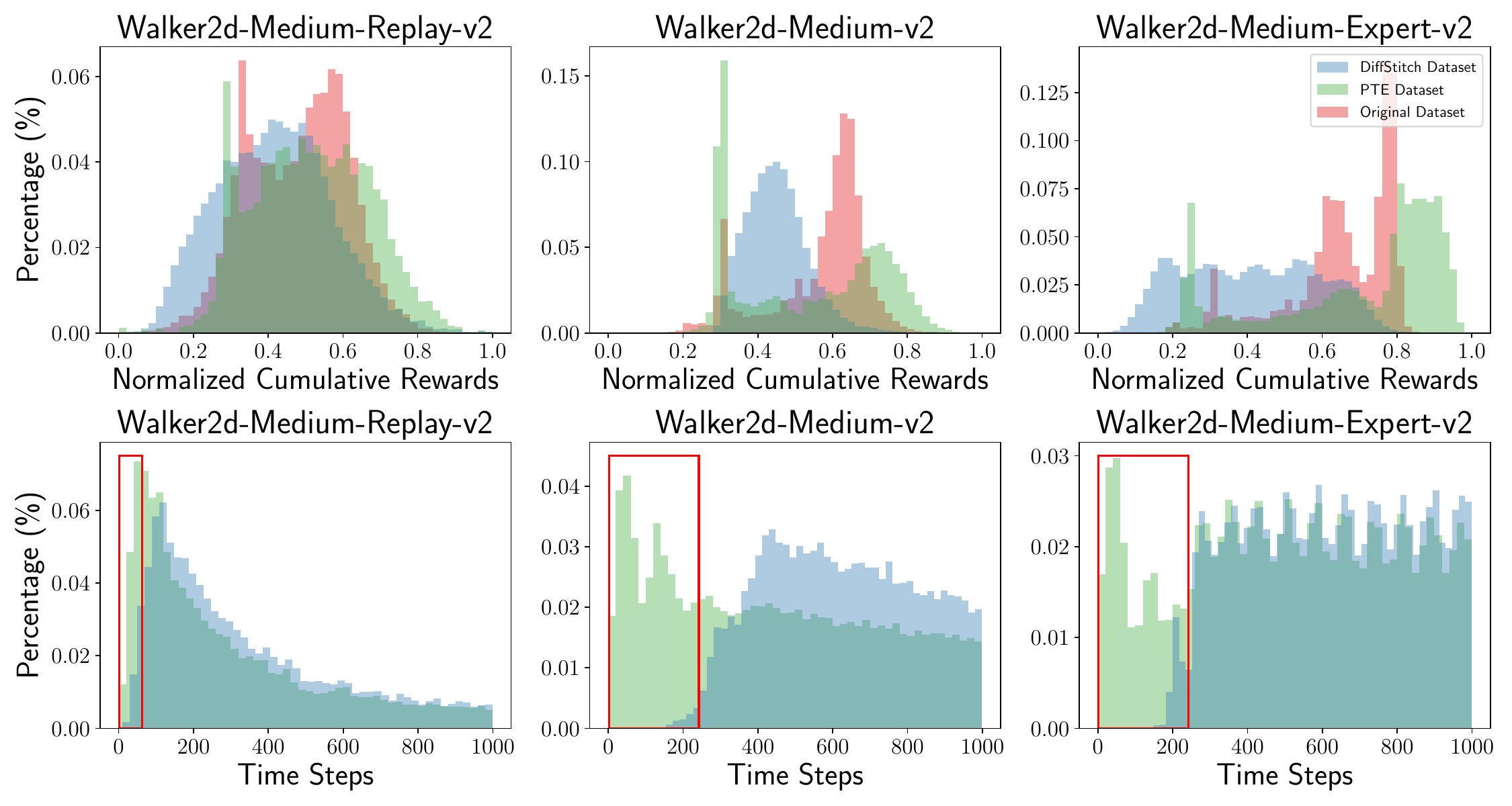}
    \caption{PTE dataset provide a wider range of distribution in terms of normalized cumulative rewards and temporal time steps.}
    \label{app:fig:pte_vs_diffstitch}
\end{figure}

\textbf{PTE and Diffusion-based Data Augmentation.}\label{app:sec:pte_vs_diffstitch}
DiffStitch \citep{diffstitch} is a data augmentation method that constructs dataset by stitching together trajectories with high cumulative rewards. However, this reward-centric strategy results in a highly skewed dataset lacking coverage across the state space. As shown in Tables \ref{app:exp:pte_diffsttich} and Figure \ref{app:fig:pte_vs_diffstitch}, this bias leads to poor performance when training planners solely on DiffStitch-augmented datasets.

\begin{table}[t!]
\small
\centering
\caption{PTE outperforms other diffusion-based data augmentation method on extendable planning problem.}
\label{app:exp:pte_diffsttich}
\begin{tabular}{llrrrr}
\toprule 
\multicolumn{2}{c}{\textbf{Environment}}        & \textbf{DiffStitch}    &\textbf{Diffuser-X}    &  \textbf{DiffuserLite-DiffStitch} & \textbf{HM-Diffuser-X}   \\ \midrule
Maze2D & Large & {99.9 +- 7.2} & {114.4 +- 4.7} & \multicolumn{1}{r}{-} & \textbf{166.9} +- 4.6 \\
Maze2D & Giant & {80.9 +- 12.1} & {114.6 +-9.1} & \multicolumn{1}{r}{-} & \textbf{177.4} +- 11.9 \\
\midrule
\multicolumn{2}{c}{\textbf{Maze2D Average}} & 90.4 & 114.5 & \multicolumn{1}{r}{-} & \textbf{172.1} \\ \midrule
Kitchen & Partial-v0 & 21.7 +- 2.1 & 43.3 +- 3.2 & 13.3 +- 3.1 & \textbf{56.7} +- 5.3 \\
Kitchen & Mixed-v0 & 18.3 +- 2.7 & 48.3 +- 4.7 & 20.0 +- 2.5 & \textbf{61.7} +- 3.1 \\ \midrule
\multicolumn{2}{c}{\textbf{Kitchen Average}} & 20 & 45.8 & 16.6 & \textbf{59.2} \\ \midrule
Walker2d & MedReplay & 34.1 +- 6.1 & 20.1 +- 4.3 & 12.2 +- 1.4 & \textbf{29.6} +- 4.8 \\
Walker2d & Medium & 31.3 +- 7.2 & 62.6 +- 4.3 & 35.9 +- 3.7 & \textbf{72.7} +- 2.5 \\
Walker2d & MedExpert & 0.0 +- 0.0 & \textbf{80.3} +- 3.7 & 17.7 +- 1.6 & 79.3 +- 2.3 \\ \midrule
\multicolumn{2}{c}{\textbf{Walker2d Average}} & 21.8 & 54.3 & 21.9 & \textbf{60.5} \\ \midrule
Hopper & MedReplay & 24.6 +- 3.8 & {34.5} +- 6.2& 20.0 +- 4.2 & \textbf{37.3} +- 4.8 \\
Hopper & Medium & 1.0 +- 0.0 & 44.3 +- 3.5 & 37.6 +- 6.4 & \textbf{44.9} +- 3.5 \\
Hopper & MedExpert & 0.0 +- 0.0 & \textbf{74.9} +- 8.0 & 17.3 +- 1.8 & 74.3 +- 9.0 \\ \midrule
\multicolumn{2}{c}{\textbf{Hopper Average}} & 8.5 & 51.2 & \multicolumn{1}{r}{24.9} & \textbf{52.1} \\ 

\bottomrule
\end{tabular}
\end{table}

\begin{figure}[ht]
    \centering
    \includegraphics[width=0.7\textwidth]{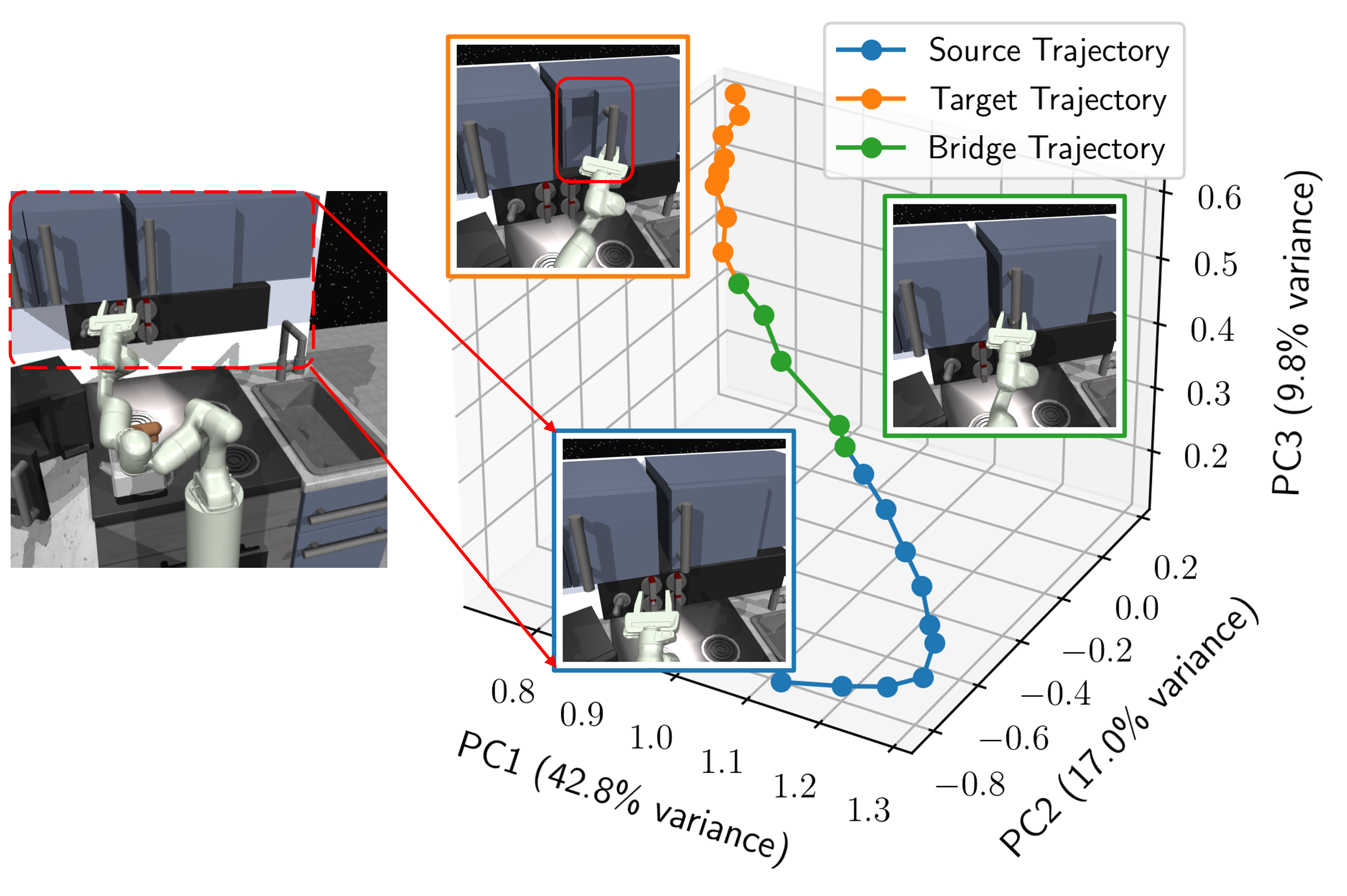}
    \caption{The bridge trajectory connects the source and target behaviors, where the robot arm completes the subtask of slide door, resulting a succefull demonstration of a subtask.}
    \label{app:fig:kitchen_stitch}
\end{figure}

\end{document}